%% file: main.tex
\documentclass[runningheads]{llncs}

\usepackage{eccv}

\usepackage{eccvabbrv}

\usepackage{graphicx}
\usepackage{booktabs}

\usepackage[accsupp]{axessibility}  

\usepackage{algorithm}
\usepackage{algpseudocode}
\usepackage{amsmath} 
\usepackage{amssymb}
\usepackage{bbding}
\usepackage{bm}
\usepackage{fontawesome5}
\usepackage{graphicx}
\usepackage{makecell}
\usepackage{multirow}
\usepackage{tabularx}
\usepackage{xcolor}
\usepackage{pifont} 
\usepackage{enumitem}
\usepackage{wrapfig}
\usepackage[title,toc,titletoc,header]{appendix}
\newcommand{\cmark}{\ding{51}}
\newcommand{\xmark}{\ding{55}}

\definecolor{cvprblue}{rgb}{0.16,0.32,0.75}
\providecolor{cvprblue}{rgb}{0.16,0.32,0.75}

\usepackage[pagebackref=true,breaklinks=true,colorlinks,
            linkcolor=cvprblue,
            citecolor=cvprblue,
            urlcolor=cvprblue,
            bookmarks=false]{hyperref}
\usepackage{pifont}

\newcommand{\OurHours}{100.69}
\newcommand{\MethodName}{ChoreoLLaMA}

\newcommand{\repeatthanks}{\textsuperscript{\thefootnote}}

\usepackage{orcidlink}

\begin{document}

\title{\textit{InfiniteDance}: Scalable  3D Dance Generation Towards in-the-wild Generalization} 

\titlerunning{Scalable 3D Dance Generation Towards in-the-wild Generalization}

\author{{Ronghui Li}\thanks{co-first authors; $^{\dagger}$ corresponding author}$^{1,2}$\quad 
{Zhongyuan Hu}\repeatthanks$^{,1}$\quad 
Li Siyao$^3$\quad 
Youliang Zhang$^1$\quad 
\\Haozhe Xie$^{3}$\quad 
Mingyuan Zhang$^{3}$\quad 
Jie Guo$^2$\quad 
Xiu Li$^{\dagger,1}$\quad 
Ziwei Liu$^{3}$
}

\authorrunning{Li, Hu et al.}

\institute{$^1$Tsinghua University\quad,
$^2$Peng Cheng Laboratory\\
$^3$S-Lab, Nanyang Technological University\\
\vspace{8pt}
Project Page: \url{https://infinitedance.github.io/}
}

\maketitle

\input{sec/0_abstract}   
\input{sec/1_intro}
\input{sec/2_related}
\input{sec/3_dataset}
\input{sec/4_method}

\input{sec/5_experiment}

%
%
\bibliographystyle{splncs04}
\bibliography{main}
\newpage

\begin{subappendices}
\renewcommand{\thesection}{~\Alph{section}}
\input{sec/X_suppl} 
\end{subappendices}
\end{document}

%% file: sec/0_abstract.tex
\begin{center}
  \vspace{-3 mm}
  \centering
  \captionsetup{type=figure}
  \vspace{-2 mm}
  \resizebox{0.85\linewidth}{!}{
    \includegraphics{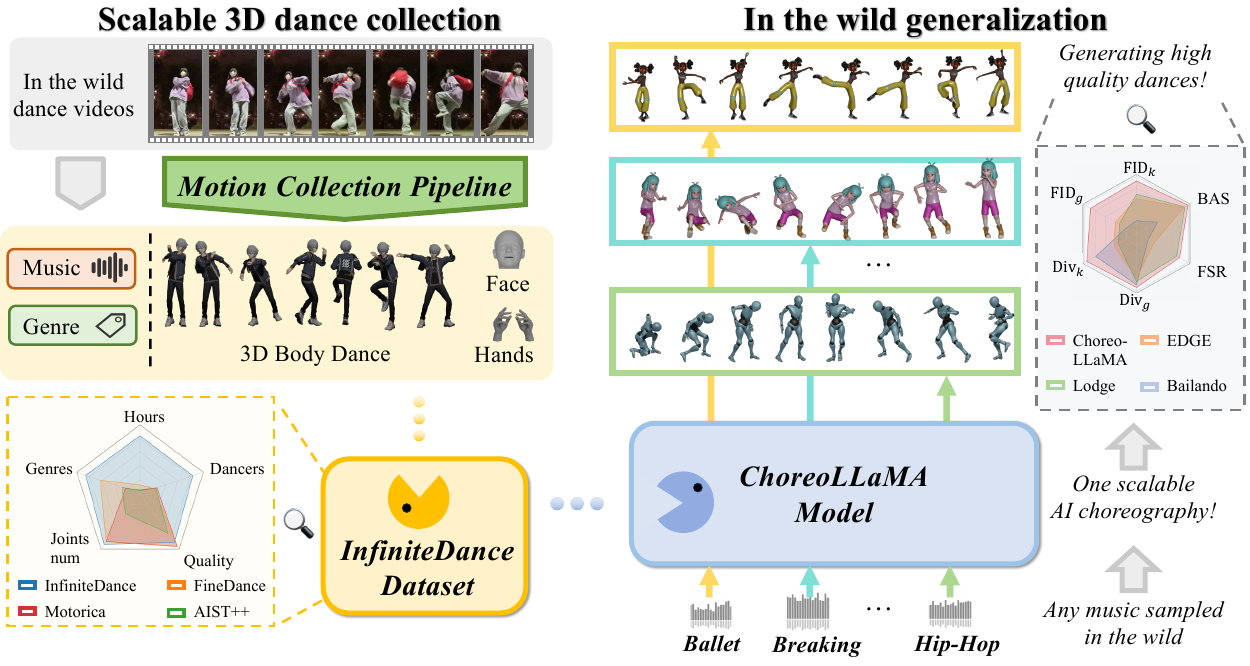}
  }
  \caption{We propose a fully automated \textbf{Motion Collection Pipeline} that extracts high-fidelity and physically plausible 3D dance motions from large amounts of in-the-wild videos. Based on this pipeline, we construct \textbf{InfiniteDance Dataset}, a 100.69 hours music–dance paired dataset.
  We further propose the \textbf{ChoreoLLaMA},
  trained on InfiniteDance, which generates high-quality 3D dances that match the tempo and style of in-the-wild music. }
  \label{fig:teaser}
\end{center}%

\begin{abstract}

Although existing 3D dance generation methods perform well in controlled scenarios, they often struggle to generalize in the wild.
When conditioned on unseen music, existing methods often produce unstructured or physically implausible dance, largely due to limited music-to-dance data and restricted model capacity. 
This work aims to push the frontier of generalizable 3D dance generation by scaling up both data and model design.
\textbf{1)} On the data side, we develop a fully automated pipeline that reconstructs high-fidelity 3D dance motions from monocular videos. 
To eliminate the physical artifacts prevalent in existing reconstruction methods, we introduce a Foot Restoration Diffusion Model (FRDM) guided by foot-contact and geometric constraints that enforce physical plausibility while preserving kinematic smoothness and expressiveness, resulting in a diverse, high-quality multimodal 3D dance dataset totaling \OurHours\ hours.
\textbf{2)} On model design, we propose Choreographic LLaMA (ChoreoLLaMA), a scalable LLaMA-based architecture. 
To enhance robustness under unfamiliar music conditions, we integrate a retrieval-augmented generation (RAG) module that injects reference dance as a prompt. 
Additionally, we design a slow/fast-cadence Mixture-of-Experts (MoE) module that enables ChoreoLLaMA to smoothly adapt motion rhythms across varying music tempos.
Extensive experiments across diverse dance genres show that our approach surpasses existing methods in both qualitative and quantitative evaluations, marking a step toward scalable, real-world 3D dance generation.
\end{abstract}

%% file: sec/1_intro.tex
\section{Introduction}
\label{sec:intro}

Generating high-quality 3D dance is essential for a wide range of applications, including 3D animation, filmmaking, digital performance, and interactive entertainment. 
%
%
In recent years, there has been increasing interest in deep learning-based methods for automatic dance generation, with the goal of developing an \textit{AI choreographer} that can significantly reduce the costly and time-consuming manual efforts involved in traditional 3D choreography pipelines, thereby enabling scalable content creation.

Despite notable progress in recent years, state-of-the-art  3D dance generation methods are still not ready to be deployed in real-world applications.
This is mainly due to two key limitations:
\textbf{(1) Dance motion quality remains suboptimal.}
Current methods still often produce basic artifacts such as foot skating and mesh penetration. 
%
%
\textbf{(2) Limited generalization across in-the-wild music conditions.}
Although many existing approaches can excel in controlled settings, they often collapse into \textit{unstructured motions} under diverse musical conditions.
On one hand, this limitation arises from the scarcity and imbalance of existing music-to-dance datasets, which fail to provide sufficient scale and diversity for models to learn the broad genres of musical and choreographic patterns needed for robust generalization.
%
On the other hand, existing methods often depend on handcrafted and biased music-conditioning designs, yielding limited adaptability to diverse musical styles and tempos. 
For example, Lodge~\cite{li2024lodge} performs reliably on fast-tempo tracks but struggles with slower rhythms, as its manually crafted rules are tailored for high-energy dance.
%

Inspired by recent breakthroughs in large-scale models, particularly in LLMs and video generation, we explore whether scaling up data and model capacity can benefit 3D dance generation.
In this work, we propose \textit{InfiniteDance}, a scalable 3D dance generation framework by jointly scaling both \emph{data} and \emph{model} capacity, with the goal of advancing deep learning-based methods toward more practical and deployable dance synthesis.

First, on the data side, we introduce an automatic pipeline that converts monocular videos into 3D dance motion and use it to construct a large-scale, high-quality dataset. 
Although MoCap-based datasets provide high-fidelity motion, their scale is limited to only a few hours. In contrast, monocular reconstruction is more scalable but frequently suffers from artifacts, including penetration, jittering, floating, and foot skating.
%
Recent motion imitation methods~\cite{luo2023perpetualhumanoidcontrolrealtime,zhang2024plug}, built on physical simulators~\cite{erez2015simulation,makoviychuk2021isaac}, improve physical plausibility but often suffer from foot jittering due to inaccurate estimation of diverse in-the-wild ground friction.
To address these issues, we propose a Foot Restoration Diffusion Model (FRDM), which repairs foot-related artifacts in a self-supervised manner using velocity, position, and rotation cues, while preserving fidelity to the original motion. With geometric guidance during inference, FRDM significantly improves motion realism and brings reconstruction quality close to professional MoCap. 
%
Based on this pipeline, we build \textit{InfiniteDance}, a dataset containing \OurHours hours of high-quality 3D dance–music pairs covering $30$ dance genres, along with RGB videos, 2D keypoints, and other annotations.

Second, on the modeling side, we propose \textit{ChoreoLLaMA}, a scalable architecture that maps music and motion conditions into learnable tokens instead of relying on handcrafted priors. 
We use a pretrained LLaMA~\cite{touvron2023LLaMAopenefficientfoundation,touvron2023LLaMA2openfoundation,grattafiori2024LLaMA3herdmodels} backbone and feed continuous token embeds extracted by MuQ (for music) and an RVQ-VAE (for dance).
To mitigate performance degradation under unseen musical conditions, we adopt a cross-modal Retrieval-Augmented Generation (RAG) strategy that selects reference dance motions paired with the music to guide generation.
Furthermore, we introduce Cadence-MoE, a Mixture of Cadence Experts designed to learn choreography behaviors across different rhythmic patterns. 
It jointly models music, genre, and dance tokens under varying tempos, while an adaptive gating network dynamically selects expert modules, enhancing style alignment and improving tempo consistency.

Powered by large-scale training on \textit{InfiniteDance}, our model produces stable and expressive dance motions, outperforming existing methods both qualitatively and quantitatively and pushing 3D dance generation closer to real-world applicability.
In conclusion, our key contributions are as follows:

\textbf{1.}  We propose a novel 3D dance collection pipeline that captures high-quality, physically plausible, and expressive motion from monocular videos. At its core is an efficient Foot Restoration Diffusion Model (FRDM) that effectively resolves foot-ground contact artifacts while preserving the geometric fidelity of the original motion.

\textbf{2.}  We construct a large-scale, high-quality 3D dance dataset, \textit{InfiniteDance}, comprising \textbf{\OurHours}~hours of motion across 30 genres, paired with rich annotations including RGB videos, 2D keypoints, music, and genre labels.

\textbf{3.}  We design a scalable LaMMA-based choreography framework that leverages retrieved reference dances to improve generalization to in-the-wild music and employs a Cadence-MoE to mitigate generation bias caused by dataset imbalance, thereby enhancing music–dance style consistency. 

%% file: sec/2_related.tex
\section{Related Works}
\subsection{Choreography Dataset}
With the rise of data-driven generative models, 3D dance datasets have become essential for choreography research. Current approaches for collecting dance motion data include marker-based or inertial motion capture, multi-view camera setups, manual keyframing, and monocular video-based pose estimation.

AIST++ \cite{aist++} marked a milestone by providing 5.2 hours of music-dance paired data; They leverage multi-view camera systems and SMPLify~\cite{loper2023smpl} to reconstruct SMPL-format motion. 
MotoricaDance~\cite{motorica2024} employed professional motion capture equipment, offering  6.2 hours of high-quality data.
FineDance~\cite{li2023finedance} further collected 14.6 hours of dance data by a marker-based MoCap system and professional dancers with fine-grained hands. 
Although using MoCap equipment can ensure motion quality, the setup is expensive and restricts the capture environments.

Danceformer~\cite{danceformer} introduced another important category of 3D dance datasets: those created using animation software through manual keyframing by professional animators.
DanceCamera3D~\cite{wang2024dancecamera3d3dcameramovement} further collected dance motions from the MMD (MikuMikuDance) community, which are typically keyframed by enthusiasts.
However, the motions remain animator-edited and often lack natural dynamics.

Recently, PoPDanceSet~\cite{luo2024popdgpopular3ddance} used monocular video-based motion capture to collect dance data from in-the-wild videos. They gathered popular dance clips online and used the HybrIK~\cite{li2021hybrik,li2025hybrik} model to estimate 3D poses, resulting in 3.56 hours of motion data. However, monocular estimation lacks physical modeling and often introduces artifacts such as jitter, mesh penetration, floating, and foot skating.

In summary, existing datasets struggle to achieve both scalability and high quality, and none of them include facial expressions. Our dataset addresses these gaps by providing 100.69 hours of high-quality 3D dance motion data across 30 diverse genres, with detailed hand movements and expressive facial annotations.

\subsection{Music Driven Dance Generation}

Generating music-synchronized dance has been studied extensively. Traditional motion-graph-based approaches~\cite{choreographers_motiongraph,berman2015kinetic,ofli2011learn2dance,ChoreMaster} retrieve candidate dance clips based on musical features and stitch them together using hand-crafted choreography rules. However, these methods struggle to generalize across dance genres due to the complexity and diversity of choreographic patterns.

With the rise of deep learning, some early works~\cite{kim2022brand_sequencemodel_based,aist++,gtn,luo2024m3gpt,zhang2025motion} treat music2dance as a seq2seq task and use LSTM or Transformer to generate frame-by-frame. However, these methods suffer from motion-freezing issues because of error accumulation. 
Diffusion-based methods~\cite{li2023finedance,edge,li2024lodge,Sun_2024,zhang2023teditemporallyentangleddiffusionlongterm,Goel_2024,cohan2024flexiblemotioninbetweeningdiffusion} model motion features in continuous space through iterative denoising. EDGE~\cite{edge} and FineDance~\cite{li2023finedance} produce high-quality short clips, and Lodge~\cite{li2024lodge} introduces a coarse-to-fine framework for long-term choreography, producing impressive results for fast-paced dances. However, its handcrafted dance primitives and rule-based choreography augmentation do not generalize well to diverse dance genres.

Token-based autoregressive  methods~\cite{siyao2022bailando,luo2024m3gpt,aist++} use discrete motion tokens for compact and high-quality motion representation and then train a sequence model to learn music-dance dependencies. Based on this, Bailando~\cite{siyao2022bailando} enhances rhythmicity via reinforcement learning.
However, it only takes low-level music features as input, making it challenging for the sequence model to model the high-level musical structure. As a result, the generated dances often lack coherent choreographic structure and may exhibit repetitive or meaningless motions, such as random hand waving without semantic intent.

Existing methods are limited to certain dance genres, resulting in poor generalization to diverse genres and unfamiliar music. Our approach supports multiple genres and improves choreography quality and generalization through a Retrieval Augmented Generation (RAG) mechanism and Cadence-MoE.



\renewcommand{\arraystretch}{0.8}
\begin{table*}[t!]
    \caption{\small{\textbf{Comparisons of 3D Dance Datasets.}  \faHandPaper[regular] denotes whether containing hand (finger) motion. \faGrinWink[regular] represents facial expression. {$\bar{\text{T}}$(sec)} denotes the average seconds per sequence. Acquisition methods: MoCap (captured with professional motion capture equipment), Pseudo (previous video-based motion estimation), Animator (manually keyframed by animators).}
    }
    \vspace{-3mm}
	
	\centering
	\resizebox{0.9\textwidth}{!}{
		\begin{tabular}{lccccccccc}
			\toprule 
			Dataset & Acquisition & \makecell[c]{Joints num} & 
            \faHandPaper[regular]
            & \makecell[c]{\faGrinWink[regular]} & Genres & Representation & Dancers & \makecell[c]{T(h)} & \makecell[c]{$\bar{\text{T}}$(s)} \\
			\hline\noalign{\smallskip}
			Dance w/. Melody\cite{dancewithmelody} &MoCap & 21 & \XSolidBrush & \XSolidBrush & 4 & 3D joints & - & 1.6 & 92.5 \\
			Music2Dance\cite{Music2Dance} & MoCap & 55 & \Checkmark & \XSolidBrush & 2 & 3D joints & 2 & 0.96 & - \\
			EA-MUD\cite{deepdance} & Pseudo & 24 & \XSolidBrush & \XSolidBrush & 4 & 3D joints & - & 0.35 & 73.8 \\
			PhantomDance\cite{danceformer} & Animator & 24 & \XSolidBrush & \XSolidBrush & 13 & SMPL& 100+ & 9.6 & 133.3 \\
			AIST++\cite{aist++} & Pseudo & 17/24 & \XSolidBrush & \XSolidBrush & 10 & SMPL & 30& 5.2 & 13.3 \\
			MMD\cite{ChoreMaster} & MoCap & 52 & \Checkmark & \XSolidBrush & 4 & FBX & - & 9.9 & - \\
			FineDance\cite{li2023finedance} & MoCap & 52 & \Checkmark & \XSolidBrush & 22 & SMPL-X & 27 & 14.6 & 152.3 \\
			POPDG\cite{luo2024popdgpopular3ddance} & Pseudo & 24 & \XSolidBrush & \XSolidBrush & 19 & SMPL& - & 3.56 & - \\
			Motorica\cite{motorica2024} & MoCap & 52 & \XSolidBrush & \XSolidBrush & 8 & BVH & - & 6.22 & - \\
			AIOZ\cite{wang2024dancecamera3d3dcameramovement} & Pseudo & 24 & \XSolidBrush & \XSolidBrush & 7 & SMPL & 4000+ & 16.7 & 37.5 \\
			DCM\cite{le2023musicdrivengroupchoreography} & Animator & 24 & \XSolidBrush & \XSolidBrush & 4 & MMD & - & 3.2 & 106.67 \\
			DD100\cite{siyao2024duolandofollowergptoffpolicy} & MoCap & 52 & \Checkmark & \XSolidBrush & 10 & SMPL-X & 10 & 1.92 & 69.3 \\
			InterDance\cite{li2024interdancereactive3ddancegeneration} & MoCap & 52 & \Checkmark & \XSolidBrush & 15 & SMPL-X & - & 3.93 & 142.7\\
			\midrule [0.5pt]
			\textbf{InfiniteDance (Ours)} & Our Pipeline & \textbf{55} & \textbf{\Checkmark} & \textbf{\Checkmark} & \textbf{30} & \textbf{SMPL-X} & 1000+ & \textbf{100.69} & 30.01 \\
            
			\bottomrule 
		\end{tabular}
	}
	\label{tab:data_com}
\end{table*}

%% file: sec/3_dataset.tex
\section{\textit{InfiniteDance} Dataset}
As shown in Table \ref{tab:data_com}
, We collect the \textit{InfiniteDance} dataset, sourced from wild video platforms, provides high-fidelity motions with complex actions and detailed hand and facial movements, spanning 6 main categories and 30 fine-grained genres with taxonomy verified by professional dancers.

\begin{figure*}[t]
  \centering
  \includegraphics[width=0.86\linewidth]{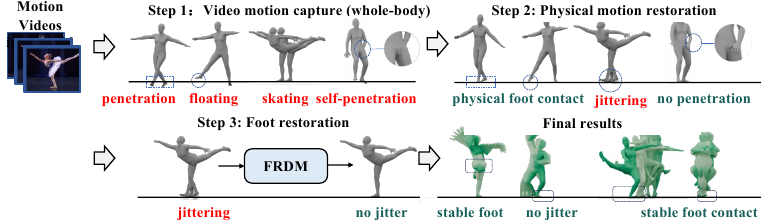}
\caption{Overview of our motion collection pipeline.
\textbf{Step 1:} We estimate whole-body motion from monocular videos, which contain artifacts.
\textbf{Step 2:} We refine these motions through motion imitation in a physics simulator to obtain more physically plausible results, but this step often introduces frequent foot jittering.
\textbf{Step 3:} We apply our Foot Restoration Diffusion Model (FRDM) to further correct foot motions.
The \textbf{final results} show stable root and foot contacts without jittering or penetration artifacts.}
\label{fig:pipeline}
\end{figure*}

As shown in Fig.\ref{fig:pipeline}, the \textit{InfiniteDance} dataset is collected using our proposed scalable and automated 3D motion collection pipeline, which extracts physically plausible motions from monocular videos.
As shown in Fig.~\ref {fig:pipeline}. The pipeline includes the following steps:

The first step is to extract high-quality whole-body motions from monocular videos using video-based motion estimation methods. We first preprocess the videos by using YOLOv8~\cite{varghese2024yolov8} to extract single-person video sequences. Given its strong generalization ability and gravity-aware modeling, we employ GVHMR~\cite{shen2024world} to estimate body motion. We use SMPLest-X~\cite{yin2025smplest} to obtain SMPL-X expression and hand parameters, as it captures visible features and estimates occluded faces and hands accurately.

The motions estimated in the previous step are used as references for motion \textit{imitation}~\cite{Luo2023PerpetualHC} within a physical simulation environment, which helps correct non-physical artifacts by enforcing physical constraints.
This step effectively eliminates common artifacts such as body interpenetration, floating, and foot skating.
However, because the physics-based simulation cannot accurately model the diverse ground–surface frictions involved in different dance movements, it often converts foot-skating artifacts into noticeable \textit{foot jittering}.

To further address the foot-jittering issue commonly observed in motion imitation, we introduce a Foot Restoration Diffusion Model (FRDM).

\subsection{Foot Restoration Diffusion Model}
\begin{figure*}[t!]
  \centering
  \includegraphics[width=0.7\linewidth]{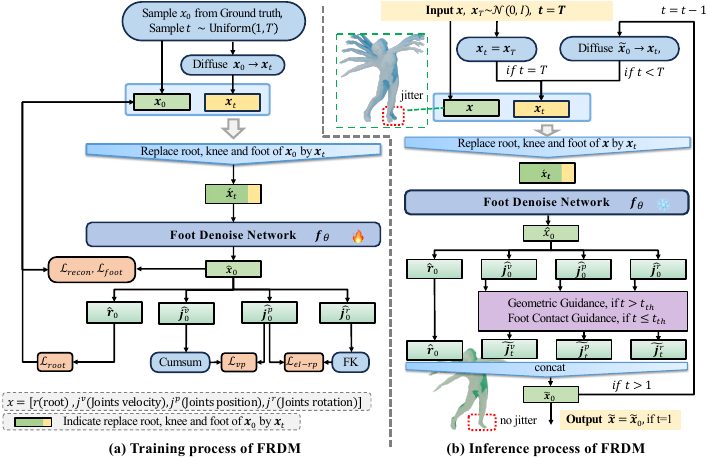}
  \caption{\textbf{(a)} The Foot Restoration Diffusion Model (FRDM) can be trained in a self-supervised manner. We sample $\bm{x}_0$ from ground-truth motions and obtain $\bm{x}_t$ by adding noise. 
  To repair only the artifacts in the root, knees, and feet, we replace these parts in $\bm{x}_0$ with the corresponding components from $\bm{x}_t$ to obtain $\acute{\bm{x}}_t$. We then train a foot denoising network $\bm{f_{\theta}}. \hat{\bm{x}}_0=\bm{f_{\theta}}(\acute{\bm{x}}_t,t)$,
  $\bm{j}^p_0=\text{Cumsum}(\bm{j}^v_0),\bm{j}^p_0=\text{FK}(\bm{j}^r_0)$.
  \textbf{(b)} Given the motion $\bm{x}$ with foot artifacts, we first sampel $\bm{x}_T \sim \mathcal{N}(0,I)$, and get $\acute{\bm{x}}_t$ by replace the root, knee and foot reigon of $\bm{x}$ by those of $\bm{x}_t$. 
  In the early denoising steps $t>t_{th}$, where $t_{th}$ is a threshold, we apply geometric guidance to keep the restored motion geometrically consistent with the original input. In the last steps $t\leq t_{th}$, we use foot-contact guidance to explicitly improve foot stability.
}
  \label{fig:footres}
\end{figure*}
Given a flawed motion sequence $\bm{x}$ with foot-ground contact artifacts, our goal is to produce a corrected motion $\tilde{\bm{x}}$ that exhibits more stable foot contacts while preserving the original full-body geometric consistency. 
Since these artifacts mainly arise from instability in the root, knees, and feet, we restrict our correction to these joints and keep the upper body unchanged.

Optimizing for foot-skating artifacts directly is difficult, as SMPL pose parameters tend to amplify small errors near the root. We therefore operate in the linear joint position space.
We adopt the motion representation similar to HumanML3D~\cite{humanml3d}.
For a motion sequence $\bm{x}\in\mathbb{R}^{L\times259}$, where $L$ is the motion frame number, each frame $\bm{x}$ can be represented as $\bm{x}=[\bm{r},\bm{j}^v,{\bm{j}}^p,\bm{j}^r]$. 
$\bm{r}=[{\dot{\bm{r}}}^a,{\dot{\bm{r}}}^x,{\dot{\bm{r}}}^z,{\bm{r}}^y]$ is the root data, ${\dot{\bm{r}}}^a$ is the angular velocity along the Y-axis (yaw angle), ${\dot{\bm{r}}}^x,{\dot{\bm{r}}}^z$ are root linear velocities on the floor, $\bm{r}^y$ is the root height. 
$\bm{j}^v\in \mathbb{R}^{L\times3J},\bm{j}^p\in \mathbb{R}^{L\times3(J-1)},\bm{j}^r\in \mathbb{R}^{L\times6(J-1)}$ correspond to the velocity, position, and rotation (smpl pose) of local keypoints relative to the root, with $J=22$ denoting the number of joints and  $J-1$ denoting all non-root joints.
They are different representations of the same motion,
Velocity aids foot stability, while Rotation helps maintain geometric consistency.
Based on this representation, we perform implicit optimization during training and apply explicit guidance during inference to correct foot artifacts while preserving the original motion as much as possible.

\subsubsection{Self-supervised Training.}

To train the FRDM for dance, we aggregate several high-quality dance datasets captured using marker-based MoCap systems, including MotoricaDance~\cite{motorica2024}, FineDance~\cite{li2023finedance}, DD100~\cite{siyao2024duolandofollowergptoffpolicy}, and InterDance~\cite{li2024interdancereactive3ddancegeneration}. All motions are retargeted to a standard SMPL-X body shape.

In classic diffusion, the model predicts the clean motion $\bm{x}_0$ from $\bm{x}_t$.
As  Fig.~\ref{fig:footres} (a) shows, the training process differs from standard diffusion pipelines. 
However, since we only aim to correct motion artifacts in the root, knee, and foot, we obtain $\acute{\bm{x}}_t$ by replacing the root, knee, and foot features in  $\hat{\bm{x}}_0$ with those of $\bm{x}_t$.
Then we denoise from $\acute{\bm{x}}_t$.
This ensures the upper body remains unchanged during denoising.

To ensure that the corrected root, knee, and foot motions remain faithful to the original sequence, we introduce several MSE-based loss terms, such as  $\mathcal{L}_{recon}(\hat{\bm{x}}_0,{\bm{x}}_0)$,$\mathcal{L}_{root}(\hat{\bm{r}}_0,{\bm{r}}_0)$. 
To encourage more stable foot–ground contact, we introduce a foot loss:
\begin{align}
\mathcal{L}_{\text{Foot}} = \sum_{{k} \in F} \left\| (\hat{\bm{P}}^{(i+1)}_k - \hat{\bm{P}}^{(i)}_{k}) \cdot \bm{b}^{(i)}_k \right\|_2^2, \quad \hat{\bm{P}} = Rec(\hat{\bm{r}}_0, \hat{\bm{j}}^p_0),
\end{align}
where $Rec(\cdot)$ is the recovery function to get global joints position $\hat{\bm{P}}$ (details in the supplementary materials), $k\in {F}$  means only select the foot joints, the $\bm{b}^{(i)}_k$ indicates the k-th foot joints whether contact with ground at frame $i$:
\begin{equation}
\bm{b}_k^{(i)} = 
\begin{cases}
1, & \text{if } \left\| \bm{P}_k^{(i+1)} - \bm{P}_k^{(i)} \right\|_2^2 < v_{\text{th}} \text{ and } h(\bm{P}_k^{(i)}) < H_{\text{th}}; \\
0, & \text{otherwise},
\end{cases}
\end{equation}
where $v_{\text{th}}$ and $h_{\text{th}}$ are velocity and height thresholds, respectively. $h(\bm{P}_k^{(i)})$ is to get the height of root form $\bm{P}_k^{(i)}$.
To constraint the generated $\hat{\bm{j}}^v_0$ and $\hat{\bm{j}}^p_0$ to be aligned, we introduce an loss $\mathcal{L}_{vp}=\left\|\text{cumsum}(\hat{\bm{j}}^v_0)-\hat{\bm{j}}^p_0 \right\|_2^2$, where $\text{cumsum}()$  integrates velocities over time to recover positions.
Notably, we expect $\hat{\bm{j}}^r_0$ and $\hat{\bm{j}}^p_0$ to remain approximately aligned, which helps balance their distinct roles during optimization: $\hat{\bm{j}}^r_0$ aims to preserve geometric consistency between the repaired and original motions, while $\hat{\bm{j}}^p_0$ emphasizes accurate foot-ground contact for artifact correction.
Therefore, we design an epsilon insensitive loss $\mathcal{L}_{\epsilon I-rp}$:
\vspace{-1.5mm}
\begin{equation}
\mathcal{L}_{\epsilon I-rp} = \sum_{k\in {KF}} \max \left( \left\| FK({\hat{\bm{j}}}^r_k) - \hat{\bm{j}}^p_k \right\|_2^2 - \epsilon, \; 0 \right)^2,
\vspace{-1.5mm}
\end{equation}
where ${k}\in {KF}$  means only select the knee and foot joints, $FK( )$ is the forward kinematic function that calculates position from rotation, $\epsilon$ is a hyperparameter that defines the allowed error tolerance.

\subsubsection{Inference.}
We argue that solely relying on a foot loss during the training phase is insufficient to fully resolve foot-ground contact issues. If the weight of foot loss is too small, it fails to sufficiently suppress foot artifacts; conversely, if set too large, it over-constrains the motion, leading to less dynamic results, thereby degrading motion expressiveness.
To address these issues, we propose Foot Contact Guidance to explicitly improve foot stability and propose Geometric Guidance to encourage geometric consistency during inference.

As illustrated in Fig.~\ref{fig:footres} (b), given a full-body motion $x$ exhibiting foot jittering and skating artifacts,
At each denoising step, the Foot Denoise Network predicts 
$\hat{\bm{x}}_0=\bm{f_\theta}(\acute{\bm{x}}_t,t)$. 
At the early denoising steps, $t > t_{th}$ , $t_{th}$ is a hyperparameter, Geometric Guidance is then applied to enforce global consistency between the restored motion and the input, formulated as:
\begin{equation}
\begin{aligned}
\hat{\bm{x}}_0 &=[\hat{\bm{r}}_0,\hat{\bm{j}}_0^v,\hat{\bm{j}}_0^p,\, \hat{\bm{j}}_0^r], \quad
\tilde{\bm{j}}_0^r = (1 - w_t)\, \bm{j}^r + w_t\,\hat{\bm{j}}^r_0, \quad w_t=t/T \\
\tilde{\bm{j}}_0^p &=\bm{b}(1 - w_t)\bm{j}^p_0 + \bm{b}\,w_t\,\hat{\bm{j}}^p_0, \quad
\tilde{\bm{j}}_0^r=\hat{\bm{j}}_0^r.
\end{aligned}
\end{equation}

At the final stage of denoising, $t < t_{th}$, we use Foot Contact Guidance to further ensure more stable foot-ground contact:
\begin{equation}
\begin{aligned}
\tilde{\bm{j}}_0^v = (1 - w_t)\, \bm{j}^v + w_t\,\hat{\bm{j}}^v_0,\quad 
\tilde{\bm{j}}_0^p = \text{cumsum}(\bm{j}^v_0),\quad 
\tilde{\bm{j}}_0^r = \hat{\bm{j}}_0^r,
\end{aligned}
\end{equation}
where $\bm{b}$ is a binary mask given by Equation~4.
We then get $\tilde{\bm{x}}_0 =[\hat{\bm{r}}_0,\tilde{\bm{j}}_0^v,\tilde{\bm{j}}_0^p,\tilde{\bm{j}}_0^r]$.
More detailed illustrations and experiments can be found at the supplementary materials.

\subsection{Quality of the InfiniteDance Dataset}

As shown in Table~\ref{tab:dataset_quality}, we use the Foot Skating Ratio (FSR)~\cite{li2024lodge}, Jitter~\cite{shen2024world} and Peneration Rate~\cite{li2024lodge++} to evaluate the motion quality.
GVHMR, the advanced video motion capture method, also struggles with in-the-wild dance videos, showing high foot-skating-rate(FSR, 28.63$\%$), jitter(31.89), and penetration(0.79$\%$). Adding a physical environment based motion imitation module (PHC~\cite{Luo2023PerpetualHC}) reduces skating and penetration but introduces fidelity loss and severe, unnatural leg jitter due to hard constraints. 
Simple smoothing reduces jitter but worsens motion fidelity, increasing skating.
In contrast, FRM uses multi-view representation, data-driven priors, and diffusion guidance to better balance fidelity and realism. The final configuration (GVHMR+PTM+FRM) achieves the lowest FSR(5.09$\%$) and jitter(14.33$\%$), comparable to the marker-based FineDance dataset (FSR 6.22$\%$, jitter 12.69$\%$),with low penetration.
FRM follows EDGE's~\cite{edge} architecture, trained for 16 hours on 2 A100, with an average inference time of 1.83s with 1024 frames.
Our dataset achieves lower FSR and Penetration Rate than the marker-based MoCap FineDance dataset~\cite{li2023finedance}, confirming competitive physical fidelity. 


\begin{table}[t]
\caption{Motion quality of InfiniteDance vs.\ marker-based MoCap. 
}
\label{tab:dataset_quality}
\centering
\resizebox{0.72\linewidth}{!}{
\begin{tabular}{lcccc}
\toprule
Method & Dataset & FSR$\downarrow$ & Jitter$\downarrow$ & Penetration$\downarrow$ \\
\midrule
MoCap & FineDance~\cite{li2023finedance}  & 6.22\% & 12.69 & 0.6954\% \\
GVHMR & - & 28.63$\%$  & 31.89& 0.7864$\%$ \\
GVHMR+PHC & - & 8.87$\%$  & 78.60& 0.0536$\%$ \\
GVHMR+PHC+Smooth & - & 14.29$\%$  & 15.39& 0.0561$\%$ \\
\textbf{GVHMR+PHC+FRDM (Ours)} & \textbf{InfiniteDance}            & \textbf{5.09\%} & 14.33 & \textbf{0.0559\%} \\
\bottomrule
\end{tabular}
}
\vspace{-6mm}
\end{table}

%% file: sec/4_method.tex
\section{Methodology}
Our goal is to generate high-quality 3D dance motions that accurately match the tempo, style, and structural rhythm of in-the-wild music. To achieve this, we design ChoreoLLaMA, a scalable choreography framework composed of two key ideas:
\textbf{(1) RAG-based Choreography:}
Instead of relying solely on music embeds, we retrieve top-k reference dances that share similar musical attributes, providing strong choreographic priors for rare or unseen music.
\textbf{(2) Cadence-MoE:}
We decompose the reference motions into multiple frequency bands and process them via specialized Experts. This design mitigates data imbalance, enables structured motion fusion, and produces expressive, multi-frequency conditioning signals.

\subsection{Tokenizer}

\begin{figure*}[t]
  \centering
  \includegraphics[width=0.7\linewidth]{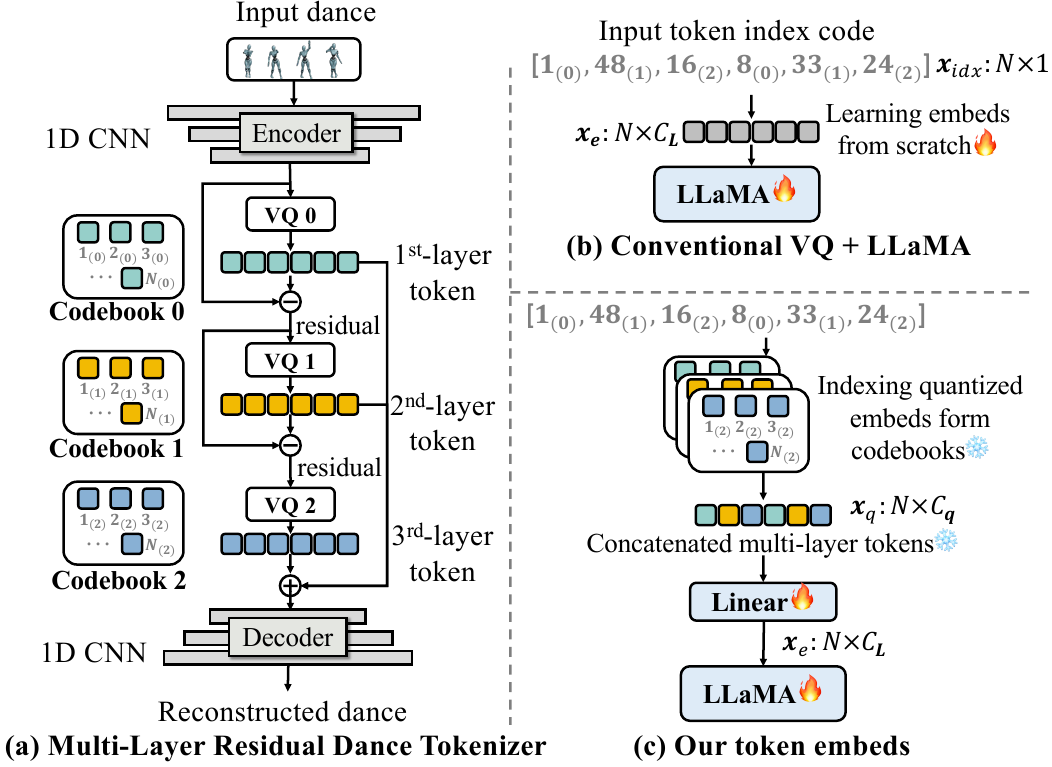}
  \caption{\textbf{(a)} Our residual tokenizer maintains multi-layer codebooks. \textbf{(b)} Previous methods input discrete indices (e.g., ``[$48_{(1)}$]'') to LLaMA. \textbf{(c)} We project continuous quantized embeds $\bm{x}_q$ into $\bm{x}_e$ for LLaMA, preserving fine-grained  features.} 
  \vspace{-3mm}
  \label{fig:Tokenizer} 
\end{figure*}

To enable LLaMA to model the cross-modal correspondence between music and dance, we first design a task-specific tokenization strategy for both modalities.
Specifically, for music tokenization, we adopt the pretrained MuQ model~\cite{zhu2025muqselfsupervisedmusicrepresentation} to extract music features $\bm{m}_f \in \mathbb{R}^{N\times C_m}$, which are then projected through a linear layer into the final music embeddings $\bm{m}_e \in \mathbb{R}^{N\times C_L}$.
As to dance tokenization, As shown in Fig.~\ref{fig:Tokenizer}, we train a dance tokenizer following RVQ-VAE~\cite{guo2024momask} to enhance motion quality and preserve fine-grained details. In the Dance Projection module, we first obtain three-layer discrete tokens $\bm{x}_{\text{idx}}$ and continuous quantized embeddings $\bm{x}_q \in \mathbb{R}^{N\times C_q}$. The quantized embeddings are then flattened and linearly projected to produce the final dance embeddings $\bm{x}_e \in \mathbb{R}^{N\times C_L}$.

Unlike prior approaches~\cite{t2mgpt,motiongpt,ling2024motionllama}, which directly feed discrete token indices for both music and motion while ignoring their embedding representations, our design explicitly preserves continuous feature information. We argue that relying solely on token indices makes it difficult for the model to learn detailed and temporally aligned dependencies between music and motion.

Given a music clip $\bm{m}$, previous methods first tokenize it into $N$ discrete units and feed only the token indices $\bm{m}_{\text{idx}} \in \mathbb{R}^{N\times 1}$ to LLaMA. LLaMA must then learn the corresponding embeddings $\bm{m}_e \in \mathbb{R}^{N\times C_L}$ from scratch, where $C_L$ denotes the embedding dimension. This inevitably discards important musical cues (especially low-level rhythmic structures), leading to misalignment between the generated fine-grained motion and the underlying music.

As shown in Fig.\ref{fig:Tokenizer}(c), ChoreoLLaMA instead operates on continuous embeddings rather than discrete token indices.
This representation is temporally compact yet rich in expressive detail, allowing ChoreoLLaMA to more effectively model both global structures and local rhythmic dependencies between music and dance.

\subsection{ChoreoLLaMA}

\begin{figure*}[t]
  \centering
  \includegraphics[width
  =0.95\linewidth]{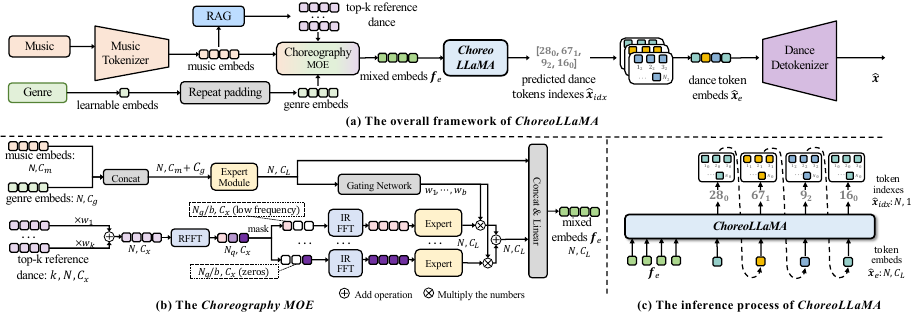}
  \vspace{-3mm}
  \caption{(a) Given a music clip and a target genre, we first use RAG to retrieve the top-k most relevant reference dances.
  These retrieved dances, together with the given music and genre embeds, are then fed into the Cadence-MoE to produce fused embeds.
  ChoreoLLaMA then autoregressively predicts dance tokens that are decoded into dance sequences.
  (b) Each ``Expert'' is a neural module composed of linear layers, multi-head attention, and a Mamba block. 
  The ``RFFT'' and ``IRFFT'' refer to the real-valued Fast Fourier Transform and its inverse, respectively. 
  (c) At the inference phase, the ChoreoLLaMA predicts dance token indices $\hat{\bm{x}}_{idx}$ one by one, we then lookup the quantized embeds $\hat{\bm{x}}_{q}$ in codebooks and project them in to dance embeds $\hat{\bm{x}}_{e}$.
  }
  \vspace{-3mm}
\label{fig:InfiniteNet}
\end{figure*}

To achieve scalable dance generation that is suitable for any given music, we propose ChoreoLLaMA, a music-driven dance generation model, as illustrated in Fig.\ref{fig:InfiniteNet}.
\noindent\textbf{RAG-based Choreography.}
To improve generalization for diverse and even rare music, we propose a Retrieval Augmented Generation (RAG) based choreography method. 
We train a Music-Dance Cross-Modal Retrieval Network (MD-Retrieval), following the CLIP~\cite{radford2021learning} architecture, where the Music Encoder and Dance Encoder utilize efficient attention~\cite{shen2021efficient}, and the model is trained using the InfoNCE~\cite{infonce} loss on the training set of the InfiniteDance dataset.

During the training and inference of ChoreoLLaMA, we retrieve the top-k most relevant training-set reference dance $\{ \bm{x}^r \}_{r=1}^{k}$. Each ${\bm{x}_r}$ is processed through a linear projection operation to obtain dance embeds \( {\bm{x}^r_e}\in \mathbb{R}^{N\times C_L} \). The final reference  embeds ${\bar{\bm{x}}^r\in \mathbb{R}^{N\times C_L}}$ is the weighted sum of $\bm{x}^r$.

\noindent\textbf{Cadence MoE.}
To capture both high-frequency motion dynamics and low-frequency, graceful movements, and to effectively leverage choreographic priors from retrieved reference dances, we propose the Cadence-MoE Network. 
For genre $\bm{g}$, we learn an embedding and repeat it to obtain $\bm{g}_e \in \mathbb{R}^{N \times C_g}$.
As shown in Fig.~\ref{fig:InfiniteNet}(b), the reference dances are weighted by $[\omega_1, \ldots, \omega_k]$, where $k_i = i / \sum_{j=1}^{k} j$, and summed to $\bar{\bm{x}}_r$.
We then apply the Real-valued Fast Fourier Transform (RFFT) to obtain frequency-domain features $\bar{\bm{x}}_f \in \mathbb{R}^{N_q \times C_x}$, where $N_q = N/2 + 1$ corresponds to the Nyquist frequency.
A frequency mask divides the spectrum into $b$ bands, each containing $(N_q/b, C_x)$ valid value and zeros elsewhere.
Each band is processed by an Expert, and their outputs are combined using weights $[\gamma_1, \ldots, \gamma_b]$ predicted by a gating network consisting of a linear layer followed by a softmax.
This design allows each expert to focus on different frequency characteristics, enabling the model to better adapt to various dance styles ranging from smooth, slow movements to fast, dynamic ones.


%% file: sec/5_experiment.tex
\section{Experiments}
\subsection{Experimental Setup}

\noindent\textbf{Datasets.} We train \MethodName\ on a dataset collection combining our InfiniteDance dataset with several public datasets\cite{aist++,li2023finedance}. InfiniteDance is split into training, validation, and test sets (85\%, 5\%, and 10\%), with genre distributions kept consistent across splits. Public datasets follow their original splits. 

\noindent\textbf{Implementation details.} 
Dataset construction used eight A100 GPUs, involved running video-based motion capture for 5 days, performing physical motion imitation for one month, and applying FRDM-based post-processing for 4 days.
Body motion was tokenized using an RVQ-VAE with 3 separate codebooks ($512$ entries, $1024$-dim), 
trained on one A100 GPU for 24 hours.
For the Foot restoration model,
We set $\epsilon$ as $0.1$, $V_{\text{th}}$ as $0.001$, set $H_{\text{th}}^{toe}$ as $0.05$, set $H_{\text{th}}^{ankle}$  as $0.08$.
For dance generation,  \MethodName\  initialized from \textit{LLaMA3.2-1B}. \MethodName\ was trained with batch size of $8$ and learning rate of $3\times10^{-4}$. 
For the RAG, we retrieve the top-10 reference dance sequences. 
For the Cadence-MoE, we divide $\bar{\bm{x}}_f$ into 2 frequency bands.
The dimension $C_m=1024,C_g=256,C_x=259,C_q=1024,C_L=2048$. 
\MethodName \ used a temperature of $0.85$, top-k sampling with $k=30$, and top-p sampling with $p=0.8$.
\subsection{Comparisons on the InfiniteDance dataset}
\begin{table*}[t]
  \caption{\textbf{Comparisons on the InfiniteDance dataset.} 
  The ``BAS'' measures the beat alignment degree between the music and dance. 
  ``Our Wins'' denotes the percentage of pairwise comparisons in which participants preferred our generated results over competing methods.
  } 
  \label{tab:sotacompare}
  \centering
  \begin{tabular}{l c c c c c c c}
    \toprule
    \multirow{2}*{Method} 
    & \multicolumn{3}{c}{Motion Quality} 
    & \multicolumn{2}{c}{Motion Diversity} 
    & \multirow{2}*{BAS$\uparrow$} 
    & \multirow{2}*{Our Wins$\uparrow$} 
    \\ 
    \cmidrule(lr){2-4} 
    \cmidrule(lr){5-6}
    & $\mathrm{FID}_k\downarrow$ 
    & $\mathrm{FID}_g\downarrow$ 
    & \makecell[c]{FSR$\downarrow$} 
    & $\mathrm{Div}_k\uparrow$ 
    & $\mathrm{Div}_g\uparrow$ 
    &  
    &  \\
    \midrule
    Ground Truth & 2.55 & 0.60 & 5.09\% & 9.37 & 7.12 & 0.2332 & $34.8 \pm 22.6\%$ \\
    \midrule
    Bailando~\cite{siyao2022bailando} & 117.38 & 82.37 & 15.56\% & 5.46 & 5.28 & 0.2137 & $88.7 \pm 8.9\%$ \\
    EDGE~\cite{edge} & 96.07 & 63.53 & 14.15\% & 4.36 & 4.97 & 0.2321 & $76.4 \pm 7.7\%$ \\
    Lodge~\cite{li2024lodge} & 89.52 & 60.38 & 6.72\% & 3.93 & 5.00 & 0.2329 & $68.1 \pm 11.5\%$ \\
    \midrule
    \textbf{\MethodName (Ours)} & \textbf{30.54} & \textbf{16.31} & \textbf{5.33\%} & \textbf{6.23} & \textbf{5.11} & \textbf{0.2342} & \textbf{-} \\
    \bottomrule
  \end{tabular}
\end{table*}

As shown in Table \ref{tab:sotacompare}, we evaluate our method in comparison with leading existing methods. To ensure a fair comparison, the results for EDGE, LODGE, and Bailando were reproduced by us on the same training dataset, following their official implementations.

\noindent\textbf{Motion Quality.} 
To evaluate the generated dance quality, we adopt the FID metric introduced in \cite{siyao2022bailando,li2024lodge}, which compares motion features between generated and ground-truth sequences using Frechet Inception Distance (\textbf{FID}) \cite{heusel2018ganstrainedtimescaleupdate}. We further evaluate foot-ground contact quality using the Foot Skating Ratio (\textbf{FSR}) \cite{li2024lodge}, which measures foot sliding during contact. Table \ref{tab:sotacompare} shows that our method yields significantly lower FID and FSR than prior works.

\noindent\textbf{Motion Diversity.}
To assess the diversity of generated dance, we follow \cite{siyao2022bailando} and compute the average pairwise Euclidean distance in motion feature space. Specifically, ${\text{Div}_k}$ reflects diversity in kinematic features, while ${\text{Div}_g}$ captures geometric variation. As shown in Table~\ref{tab:sotacompare}, our \MethodName ~achieves the highest ${\text{Div}_k}$ score, indicating richer variation in joint dynamics and motion patterns. Although the ${\text{Div}_g}$ score is slightly lower than Bailando’s, this may result from our emphasis on motion plausibility and temporal consistency, which can limit spatial variation. In contrast, Bailando’s higher foot skating ratio may inflate ${\text{Div}_g}$ by introducing unintended spatial variation.

\noindent\textbf{Beat Alignment Score (BAS).} 
We evaluate dance-music alignment using BAS \cite{aist++}. Our method achieves the best BAS of 0.2342. 

\noindent\textbf{User study.}
We conducted a user study where 50 participants viewed 40 random video pairs. Each pair consists of two
dance sequences: one created by our method and the
other by different methods or ground truth. We report our method's win rate in Table~\ref{tab:sotacompare}. More user study results are in the supplementary material.


\subsection{Generalization to In-the-Wild Music}
\label{sec:generalization}
We train models on InfiniteDance and test them under cross-dataset and out-of-distribution (OOD) settings.
For cross-dataset evaluation, we use AIST++~\cite{aist++} and FineDance~\cite{li2023finedance}, which differ substantially from InfiniteDance in capture setups, choreography styles, and music distributions.
For OOD evaluation, we curate an unseen-music set with BPMs outside the InfiniteDance training range, featuring rare instruments and styles (\eg, theremin, ambient, body percussion), introducing pronounced distribution shifts.
As shown in Table~\ref{tab:generalization}, \MethodName\ consistently outperforms Lodge~\cite{li2024lodge} across all settings, demonstrating stronger cross-dataset and OOD generalization.

\begin{table}[h]
\vspace{-3mm}
\caption{\textbf{Generalization experiments.}Both models are trained on InfiniteDance and evaluated on cross-dataset and OOD settings.}
\label{tab:generalization}
\centering
\resizebox{0.8\textwidth}{!}{
\begin{tabular}{llccc}
\toprule
Method & Test Setting & FID$_k$ $\downarrow$ & Div$_k$ $\uparrow$ & BAS $\uparrow$ \\
\midrule
Lodge~\cite{li2024lodge} & AIST++~\cite{aist++} (cross-dataset) & 48.73 & 4.26 & 0.2364 \\
\textbf{\MethodName\ (Ours)} & AIST++~\cite{aist++} (cross-dataset) & \textbf{35.45} & \textbf{5.79} & \textbf{0.2378} \\
\midrule
Lodge~\cite{li2024lodge} & FineDance~\cite{li2023finedance} (cross-dataset) & 106.85 & 4.14 & 0.2317 \\
\textbf{\MethodName\ (Ours)} & FineDance~\cite{li2023finedance} (cross-dataset) & \textbf{59.38} & \textbf{5.81} & \textbf{0.2382} \\
\midrule
Lodge~\cite{li2024lodge} & Unseen Music (OOD) & 119.66 & 5.13 & \textbf{0.2332} \\
\textbf{\MethodName\ (Ours)} & Unseen Music (OOD) & \textbf{56.22} & \textbf{5.52} & 0.2315 \\
\bottomrule
\end{tabular}
}
\vspace{-3mm}
\end{table}

\begin{table}[h]
\vspace{3mm}
\caption{\textbf{Ablation study of different components.} We progressively add each component to \MethodName\ and evaluate on the InfiniteDance test set.}
\label{tab:component_ablation}
\centering
\resizebox{0.68\textwidth}{!}{
    \begin{tabular}{c c c c c c c}
    \toprule
    \multicolumn{4}{c}{\textbf{Components}} & \multicolumn{3}{c}{\textbf{Metrics}} \\
    \cmidrule(lr){1-4} \cmidrule(lr){5-7}
    Token Indices & Token Embeds & RAG & MoE
    & $\mathrm{FID}_k\downarrow$ & $\mathrm{Div}_k\uparrow$ & BAS $\uparrow$ \\
    \midrule

    \cmark & \xmark & \xmark & \xmark
    & 79.84 & \textbf{13.09} & 0.2073 \\

    \xmark & \cmark & \xmark & \xmark
    & 62.87 & 5.49 & 0.2269 \\

    \xmark & \cmark & \cmark & \xmark
    & 38.74 & 6.16 & 0.2325 \\

    \xmark & \cmark & \cmark & \cmark
    & {33.14} & {6.11} & \textbf{0.2348} \\

    \xmark & \cmark & \cmark & \cmark
    & \textbf{30.54} & \textbf{6.23} & {0.2342} \\

    \bottomrule
    \end{tabular}
}
\vspace{-4mm}
\end{table}

\subsection{Ablation Studies}
\noindent\textbf{Token Embeds Inputs.}
We compare two variants of LLaMA input: (i) token indices, where the raw music and dance token indices are directly fed into LLaMA, and (ii) embedded inputs directly, where music and dance embeds are extracted by MuQ and the Dance Tokenizer through their respective projection layers.
Both variants are evaluated without incorporating the reference dance or the Cadence-MoE module.
As shown in Table~\ref{tab:component_ablation}, although directly using token indices leads to higher diversity, it significantly degrades motion quality, particularly in BAS performance, as low-level details are lost. In contrast, the embeds-based input yields better motion fidelity and beat alignment.

\noindent\textbf{RAG based Choreography}.
As shown in Table~\ref{tab:component_ablation}, 
when adding reference dance, even without the Cadence-MoE, the reference dance embeds are extracted by multi-head attention layers.
The overall performance rises sharply, 
confirming that reference dance priors enhance the generation of natural, diverse, and rhythm-aligned dances.

\noindent\textbf{Cadence-MoE}.
As shown in rows 3--4 of Table~\ref{tab:component_ablation}, incorporating the Cadence-MoE further improves the metrics. This improvement can be attributed to the MoE's ability to alleviate generation bias by assigning different experts for various dance patterns and frequency bands. Consequently, the model gains a stronger capacity to generate diverse choreography that better aligns with different musical styles and tempos, leading to higher-quality and more rhythmically coherent dances.


\section{Conclusion and Limitation}
In this work, we present a scalable framework for 3D dance generation that advances both data acquisition and model design. We introduce a 3D motion acquisition pipeline that efficiently captures large-scale, high-quality dance motions. The resulting InfiniteDance dataset provides a strong foundation for training more generalizable AI choreography models. Our proposed ChoreoLLaMA further enhances dance quality and generalization through the RAG-based Choreography and Cadence-MoE.
However, human choreography is inherently an iterative and interactive creative process, where artists refine movements through continuous experimentation, feedback, and collaboration. In contrast, ChoreoLLaMA currently produces dance sequences in a single forward pass conditioned only on music and style, without the ability to incorporate intermediate feedback. As a result, it does not yet support interactive refinement or co-creative choreography with human dancers.


%% file: sec/X_suppl.tex



\section*{A. Details of the Dataset}
\subsection*{A.1. Details of {InfiniteDance}}

\begin{figure}[b!]
  \centering
  \includegraphics[width=0.8\linewidth]{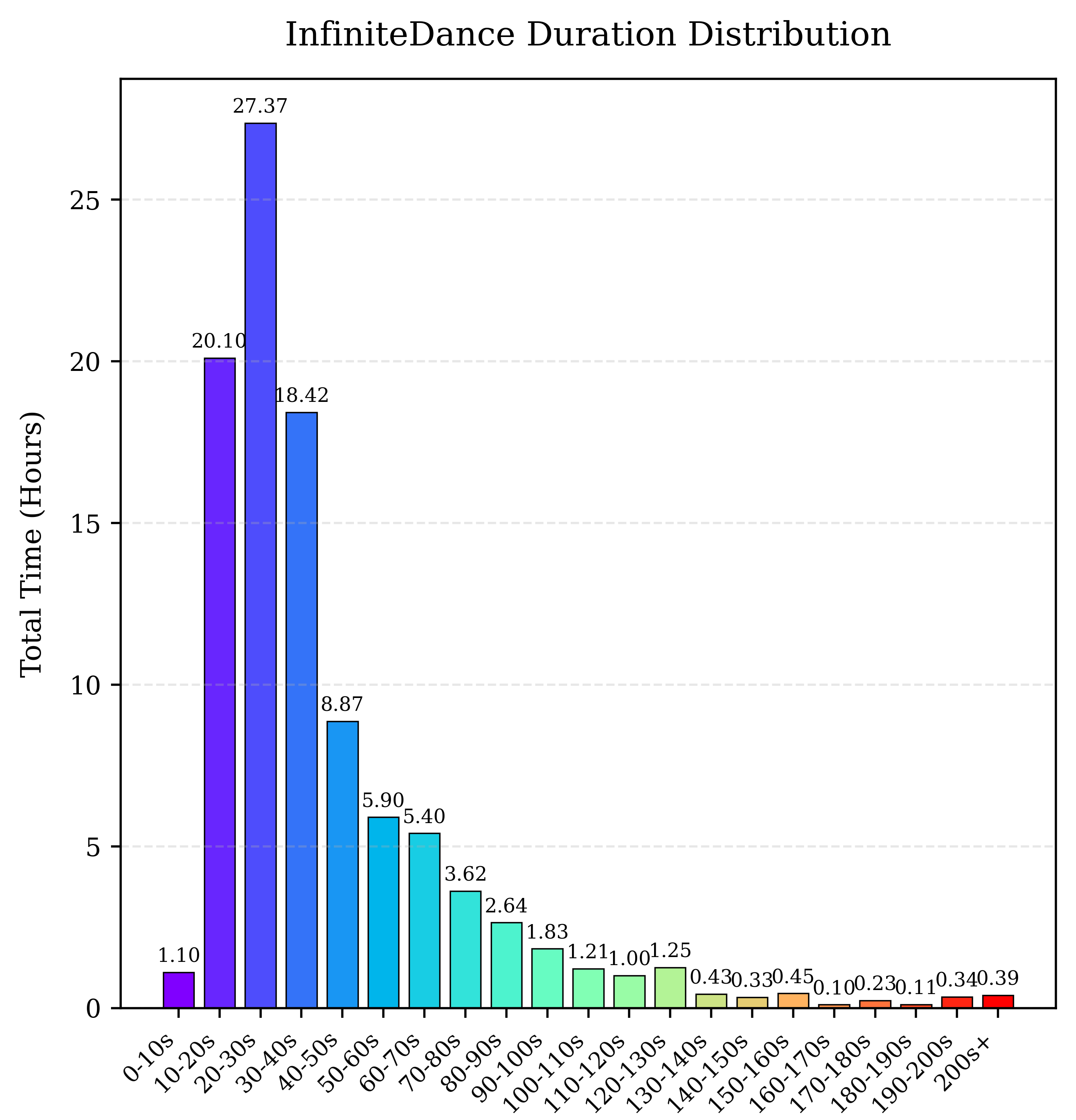}
\caption{Duration Distribution of InfiniteDance}
\label{fig:Duration}
\end{figure}

Table~\ref{tab:infinitedance_genre} and Fig.~\ref{fig:Duration} illustrate the genre and duration distribution of the {InfiniteDance} dataset, highlighting its diversity in both content and temporal span. Subcategories are denoted with hyphens (e.g., "genre-subgenre"), while "-mix" entries refer to unsegmented durations under major genres.

\begin{table}[!t]
\centering
\caption{Coarse classes and fine-grianed genres distribution of the InfiniteDance dataset.}
\label{tab:infinitedance_genre}
\resizebox{0.5\linewidth}{!}{
\begin{tabular}{llc}
\toprule
\textbf{Classes} & \textbf{Genres} & \textbf{Duration (h)} \\
\midrule
Ballet      & Ballet      & 8.31 \\
Modern      & Modern      & 3.77 \\
\midrule
Folk        & Mix         & 0.94 \\
            & Dai         & 1.48 \\
            & Uygurs      & 2.48 \\
            & Mongol      & 1.34 \\
            & Korea       & 0.46 \\
            & Zang        & 0.74 \\
            & Hehai Yangge & 0.10 \\
            & Northeastern Yangge & 0.54 \\
            & Jiaozhou Yangge & 0.48 \\
            & Chinese      & 0.06 \\
            & Miao         & 0.02 \\
\midrule
Popular     & Popular-Mix             & 5.03 \\
            & Locking  & 3.09 \\
            & Kpop            & 13.35 \\
            & HomeDance       & 6.87 \\
            & Jazz            & 7.93 \\
            & HipHop   & 9.60 \\
            & Choreography    & 6.50 \\
            & Popping- & 0.77 \\
            & Breaking & 0.17 \\
\midrule
Classic     & Classic-Mix     & 8.93 \\
            & Dunhuang & 1.27 \\
            & Shenyun  & 0.84 \\
            & Han-Tang  & 1.03 \\
            & Kun      & 0.54 \\
\midrule
Latin       & Latin-Mix       & 2.37 \\
            & Chacha    & 0.76 \\
            & Rumba     & 0.76 \\
            & Samba     & 0.51 \\
            & PasoDoble & 0.21 \\
            & Jive      & 0.34 \\
\midrule
\multicolumn{2}{r}{\textbf{Total Duration}} & \textbf{100.69} \\
\bottomrule
\end{tabular}
}
\end{table}

\subsection*{A.2. Features of InfiniteDance Dataset}
We use the above pipeline to get high-quality 3D dance, after manual verification, we obtained the InfiniteDance dataset, which has the following characteristics.

\noindent\textbf{Video Sources:}
We collect dance videos from platforms such as YouTube, TikTok, Bilibili, etc. 
These include dance tutorials and professional performance recordings. We prioritize videos that are captured with stable cameras, contain fully visible dancers, and have minimal scene cuts.

\noindent\textbf{Diverse Genre Coverage:} Our dataset includes 6 major dance genres and over 30 fine-grained subcategories, such as Ballet, Folk (e.g., Dai, Uygurs, Mongol), Popular (e.g., K-pop, Jazz, HipHop), and Classic (e.g., Dunhuang, Shen Yun). Notably, it is the first large-scale dance dataset to feature Ballet, which is underrepresented in prior works~\cite{li2023finedance,motorica2024,aist++}.Addtionally,the genre taxonomy was reviewed and verified by professional dancers.

\noindent\textbf{High-Quality Motion Data:} We obtained a high-quality InfiniteDance dataset through our carefully designed pipeline. After applying physical motion restoration and foot restoration, our sequences achieve lower FSR (5.09\%) and Penetration (0.0559\%) compared to marker-based MoCap (FineDance~\cite{li2023finedance}: FSR 6.22\%, Penetration 0.6954\%), as detailed in Section~B.3 and in the main paper. The dataset also preserves hand and facial movements, which are often omitted or poorly captured in existing datasets~\cite{aist++,li2023finedance}.

\noindent\textbf{Rich and Complex Movements:} The dataset not only features a wide range of popular dance styles such as choreography, jazz, and K-pop, but also includes technically demanding movements such as high leg lifts, pirouettes, floorwork, and single-leg spins, reflecting both artistic richness and technical diversity.

\noindent\textbf{Scale and Duration:} Since the majority of the videos are sourced from mainstream short-video platforms, the genre distribution within the dataset naturally mirrors audience preferences on these platforms. Correspondingly, the clips vary in length, ranging from a minimum duration of 6 seconds to a maximum of 4 minutes, with an average duration of approximately 29 seconds.








\section*{B. Details of FRDM}
\section*{B.1. Training Details of FRDM}

We train FRDM on a high-quality dance dataset captured using an optical motion capture system. Without requiring paired motion data with artifacts, our FRDM can be trained in a self-supervised manner on the clean motion dataset.
As shown in Alogrithm~\ref{algTrain}, at each training iteration, we sample $\bm{x}_0$ from the high-quality motion dataset, and sample $t\sim \text{Uniform}(1,\cdots, T)$, $\epsilon \sim \mathcal{N}(\mathbf{0}, \mathbf{I})$.
Then we add noise to $\bm{x}_0$ and get $\bm{x}_t$ by $\bm{x}_t = \sqrt{\bar{\alpha}_t}\bm{x}_0 +
\sqrt{1 - \bar{\alpha}_t}\boldsymbol{\epsilon}$, where $\bar{\alpha}_t$ is a noise schedule, $\bar{\alpha}_t\rightarrow 0$ when $t\rightarrow T$.
Since we only aim to fix foot-related artifacts such as foot skating and foot jitter, which are reflected in the motion of the root, knees, and feet.
Therefore, we only denoise these specific parts while keeping the rest of the motion unchanged.
Accordingly, we implement Merge by replacing the root, knee, and foot in $\bm{x}_0$ with those from $\bm{x}_t$, corresponding to Line 8, Alogrithm~\ref{algTrain}, $\acute{\bm{x}}_t \leftarrow \text{Merge}(\bm{x}_t,\bm{x}_0)$.
The we use a learnable neural network to denoise, $\hat{\bm{x}}_0=\bm{f_{\theta}}(\acute{\bm{x}}_0,t)$. Finally, we use MSE based losses $\mathcal{L}_{recon}(\hat{\bm{x}}_0,{\bm{x}}_0)$,$\mathcal{L}_{root}(\hat{\bm{r}}_0,{\bm{r}}_0)$ to ensure that the corrected root, knee, and foot motions remain faithful to the original sequence, use foot loss $\mathcal{L}_{\text{Foot}}(\hat{\bm{r}}_0,\hat{\bm{j}}_0^p)$ to implicitly enhance the foot stability. 
As shown in Equations (2) and (3) of the main paper, the foot loss $\mathcal{L}_{\text{Foot}}$ is computed based on the global joint positions $\bm{P}$, where $\bm{P}=\text{Rec}(\hat{\bm{r}}_0,\hat{\bm{j}}_0^p)$.
We also use 

Next, we explain the computation details of $\text{Rec}({\bm{r}},{\bm{j}}^p)$, $\bm{r}=[{\dot{\bm{r}}}^a,{\dot{\bm{r}}}^x,{\dot{\bm{r}}}^z,{\bm{r}}^y]$ is the root data, ${\dot{\bm{r}}}^a$ is the angular velocity along the Y-axis (yaw angle), ${\dot{\bm{r}}}^x,{\dot{\bm{r}}}^z$ are root linear velocities on the floor, $\bm{r}^y$ is the root height. We first integrate over time to obtain $[\bm{r}^a,\bm{r}^x,\bm{r}^z]$. 
Subsequently, $\bm{j}^p$ is rotated by $\bm{r}^a$ around the $y$-axis, after which the root translation $[\bm{r}^a,\bm{r}^x,\bm{r}^z]$ is added, resulting in $\bm{P}$.

\begin{algorithm}[t]
\caption{FRDM Training Algorithm}
\label{algTrain}
\begin{algorithmic}[1]
\State \textbf{Input:} Training data $\bm{x}_0 \sim q(\bm{x}_0)$, noise schedule $\beta_1, \dots, \beta_T$, $\bar{\alpha}_t = \prod_{s=1}^t \alpha_s$, $\alpha_s = 1-\beta_s$ 
\State \textbf{Output:} Trained Foot Denoise Network $\bm{f}_\theta(\acute{\bm{x}}_t, t)$
\For{Training Iterations}
\State $\bm{x}_0 \sim q(\bm{x}_0)$
\State $t \sim \text{Uniform}({1, \dots, T})$
\State $\epsilon \sim \mathcal{N}(\mathbf{0}, \mathbf{I})$

\Statex \hspace{1.1em} $\triangleright$ Diffuse $\bm{x}_0$ to $\bm{x}_t$;
$\bar{\alpha}_t \to 0$ when $t \to T$
\State $\bm{x}_t = \sqrt{\bar{\alpha}_t}\bm{x}_0 +
\sqrt{1 - \bar{\alpha}_t}\boldsymbol{\epsilon}$ 

\Statex \hspace{1.1em} $\triangleright$ Merge means replace root, knee, and foot features in $\bm{x}_0$ with those from $\bm{x}_t$;

\State $\acute{\bm{x}}_t \leftarrow \text{Merge}(\bm{x}_t,\bm{x}_0)$

\State $\hat{\bm{x}}_0 \leftarrow \bm{f}_\theta(\acute{\bm{x}}_t,t)$

\Statex \hspace{1.1em} $\triangleright$ Compute losses

\State ${\bm{x}}_0 = [{\bm{r}}_0,{\bm{j}}^v_0,{{\bm{j}}}_0^p,{\bm{j}}^r_0] $

\State $\hat{\bm{x}}_0 = [\hat{\bm{r}}_0,\hat{\bm{j}}^v_0,\hat{{\bm{j}}}_0^p,\hat{\bm{j}}^r_0] $

\State $\mathcal{L}_{recon}(\hat{\bm{x}}_0,\bm{x}_0),\mathcal{L}_{root}(\hat{\bm{r}}_0,\bm{r}_0)$ 

\State  $\mathcal{L}_{foot}(\hat{\bm{r}}_0,\hat{\bm{j}}^p_0)$
\Comment{Reduce foot artifacts.}

\State $\mathcal{L}_{vel}(\hat{\bm{j}}^v_0,\hat{\bm{j}}^p_0),
\mathcal{L}_{\epsilon I-rp}(\hat{\bm{j}}^r_0,\hat{\bm{j}}^p_0)$
\Comment{Regularize the $\hat{\bm{j}}^v_0$, $\hat{\bm{j}}^p_0$, and $\hat{\bm{j}}^r_0$ to facilitate subsequent Geometric and Foot Vel Guidance in the inference phase.}

\State Update $\bm{f}_\theta$ parameters
\EndFor

\State \Return $\bm{f}_{\theta}$
\end{algorithmic}
\end{algorithm}

\section*{B.2. Inference Details of FRDM}
After training the foot denoising network, given a motion sequence $\bm{x}$ with foot artifacts, we can obtain the corrected version $\tilde{\bm{x}}$.
As shown in Algorithm~\ref{algInfer}, we first sample $\bm{x}_T\sim \mathcal{N}(\mathbf{0},\mathbf{I})$.
The diffusion time step $t$ is from $T$ down to $1$ during the denoising process.
To ensure the upper body unchanged, we also get $\acute{\bm{x}}_t \leftarrow \text{Merge}(\bm{x}_t,\bm{x})$.
Instead predict the eposion, our framework direclty  predict the clean motion $\hat{\bm{x}}_0$ by the foot denoise network $\bm{f_{\theta}}$.

Then, we do diffusion guidance at $\hat{\bm{x}}_0$. 
As illustrated in Algorithm~\ref{algInfer}, geometric guidance is applied in the early stages of denoising to ensure that the corrected motion remains geometrically consistent with the original motion.
In the last stages of denoising, foot contact guidance is employed to explicitly address foot sliding artifacts.

\begin{algorithm}[t]
\caption{FRDM Inference Algorithm}
\label{algInfer}
\begin{algorithmic}[1]
\State \textbf{Input:} Trained Foot Denoise Network $\bm{f}_\theta$, full-body motion $\bm{x}$ with foot artifacts, hyperparameter $t_{th}$.
\State \textbf{Output:} Restored motion $\tilde{\bm{x}}$ 

\State Sample $\bm{x}_T \sim \mathcal{N}(\mathbf{0}, \mathbf{I})$
\For{$t \leftarrow T$ down to $1$}

\State $\acute{\bm{x}}_t \leftarrow \text{Merge}(\bm{x}_t,\bm{x})$  

\State $\hat{\bm{x}}_0 \leftarrow \bm{f}_\theta({\acute{\bm{x}}_t, t})$


\State ${\bm{x}} = [{\bm{r}},{\bm{j}}^v,{{\bm{j}}}^p,{\bm{j}}^r], \hat{\bm{x}}_0 = [\hat{\bm{r}}_0,\hat{\bm{j}}^v_0,\hat{{\bm{j}}}_0^p,\hat{\bm{j}}^r_0] $

\State $w_t \leftarrow t/T$

\If{$t \ge t_{th}$} \Comment{Geometric Guidance}
\State $\tilde{\bm{j}}_0^r \leftarrow (1-w_t) {\bm{j}}^r + w_t \hat{\bm{j}}^r_0$
\State $\tilde{\bm{j}}_0^p \leftarrow (1-w_t) \bm{j}^p + w_t \hat{\bm{j}}^p_0 $
\State $\tilde{\bm{j}}_0^v \leftarrow \hat{\bm{j}}_0^v $

\ElsIf{$t < t_{th}$} \Comment{Foot Vel Guidance}

\State $\tilde{\bm{j}}_{0}^v \leftarrow w_t\, \bm{b}\, \hat{\bm{j}}_{0}^v + (1 - b) \hat{\bm{j}}_{0}^v$

\State $\tilde{\bm{j}}_{0}^p \leftarrow \text{cumsum}(\hat{\bm{j}}_{0}^v)$

\State $\tilde{\bm{j}}_0^r \leftarrow \hat{\bm{j}}_0^r $

\EndIf

\State $\tilde{\bm{x}}_0 = [\hat{\bm{r}}_0,\tilde{\bm{j}}^v_0,\tilde{{\bm{j}}}_0^p,\tilde{\bm{j}}^r_0]$

\State $\bm{x}_{t-1} \leftarrow\sqrt{\bar{\alpha}_{t-1}} \tilde{\bm{x}}_0 + \sqrt{1 - \bar{\alpha}_{t-1}} \bm{x}_T$

\State $\bm{x}_{t} \leftarrow  \bm{x}_{t-1}$  

\EndFor
\State \Return $\tilde{\mathbf{x}} \leftarrow \tilde{\mathbf{x}}_0$
\end{algorithmic}
\end{algorithm}

\section*{C. More Experiments}
\section*{C.1 RAG Module Analysis}

\noindent\textbf{Retrieval Sensitivity.}
We analyze the effect of reference quality by varying the similarity rank of retrieved dances, from Top-10 (most similar) to Top-1000--1010 (least similar). As shown in Table~\ref{tab:rag_sensitivity}, even low-similarity references consistently outperform the no-RAG baseline across all metrics, demonstrating that the model is robustly benefited by retrieved dances regardless of their exact similarity rank.
\begin{table}[H]
\caption{Ablation on different retrieved reference similarity ranks.}
\label{tab:rag_sensitivity}
\centering
\resizebox{0.65\linewidth}{!}{
\begin{tabular}{lcccc}
\hline
Reference Dances & FID$_k$ $\downarrow$ & FID$_m$ $\downarrow$ & Div$_k$ $\uparrow$ & BAS $\uparrow$ \\
\hline
Top-10         & \textbf{38.74} & \textbf{13.48} & \textbf{6.92} & \textbf{0.2340} \\
Top-100--110   & 47.16 & 17.34 & 6.56 & 0.2339 \\
Top-500--510   & 51.76 & 18.73 & 6.58 & 0.2335 \\
Top-1000--1010 & 58.97 & 18.92 & 6.39 & 0.2327 \\
No RAG         & 62.87 & 147.41 & 4.68 & 0.2269 \\
\hline
\end{tabular}
}
\end{table}

\begin{figure}[h]
\centering
\includegraphics[width=0.8\linewidth]{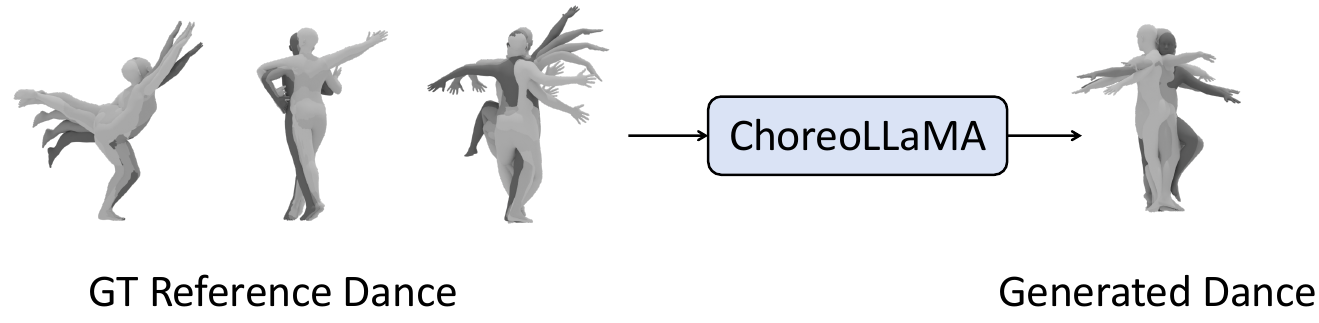}
\caption{A case show the difference between reference dance and generated dance.}
\label{fig:ref_gen_comp}
\end{figure}

\noindent\textbf{Diferences between reference and generated dance.}
We measure feature-space distances among generated motions, retrieved references, and ground-truth (GT) motions to evaluate whether the model simply copies retrieved references.
As shown in Table~\ref{tab:rag_copy}, the distance between generated and retrieved motions is substantially larger than inter-reference distances, yet generated motions remain close to the global GT distribution. This indicates that the model is influenced by retrieved choreographic priors without copying them.
To further illustrate this, Fig.~\ref{fig:ref_gen_comp} visualizes the first three frames of a reference dance alongside the corresponding generated motion. The generated poses are visibly distinct from the reference frames, providing qualitative confirmation that the model synthesizes novel motions influenced by, but not identical to, the retrieved priors.

\begin{table}[H]
\caption{Feature-space distance analysis: generated motions show influence without copying.}
\label{tab:rag_copy}
\centering
\resizebox{0.8\linewidth}{!}{
\begin{tabular}{lccc}
\hline
Distance Type & Kinematic (K) & Geometric (G) & Joint Pos. (J) \\
\hline
Retrieved--Retrieved & 11.67 & 7.97  & 27.39 \\
Generated--Retrieved & 23.52 & 10.11 & 43.13 \\
Generated--GT (avg.) & 297.94 & 20.87 & 57.11 \\
\hline
\end{tabular}
}
\end{table}

\section*{C.2 Long-term Choreography Evaluation}

We test long-sequence generation on sequences well beyond 30 seconds, observing stable FID and diversity without collapse. As shown in Table~\ref{tab:longterm}, \MethodName\ significantly outperforms Lodge~\cite{li2024lodge} in long-sequence settings, demonstrating coherent long-term choreography without quality degradation.

\begin{table}[H]
\caption{Long-term choreography evaluation (sequences $>$ 30 seconds).}
\label{tab:longterm}
\centering
\resizebox{0.72\linewidth}{!}{
\begin{tabular}{lccccc}
\hline
Method & FID$_k$ $\downarrow$ & FID$_m$ $\downarrow$ & Div$_k$ $\uparrow$ & Div$_m$ $\uparrow$ & BAS $\uparrow$ \\
\hline
Lodge~\cite{li2024lodge}      & 106.85 & 91.30 & 4.14 & 4.43 & 0.2306 \\
\MethodName\ (Ours) & \textbf{39.72} & \textbf{24.45} & \textbf{6.01} & \textbf{4.38} & \textbf{0.2335} \\
\hline
\end{tabular}
}
\end{table}

\section*{C.3 Cadence-MoE Frequency Band Analysis}

\begin{figure}[t]
\centering
\includegraphics[width=\linewidth]{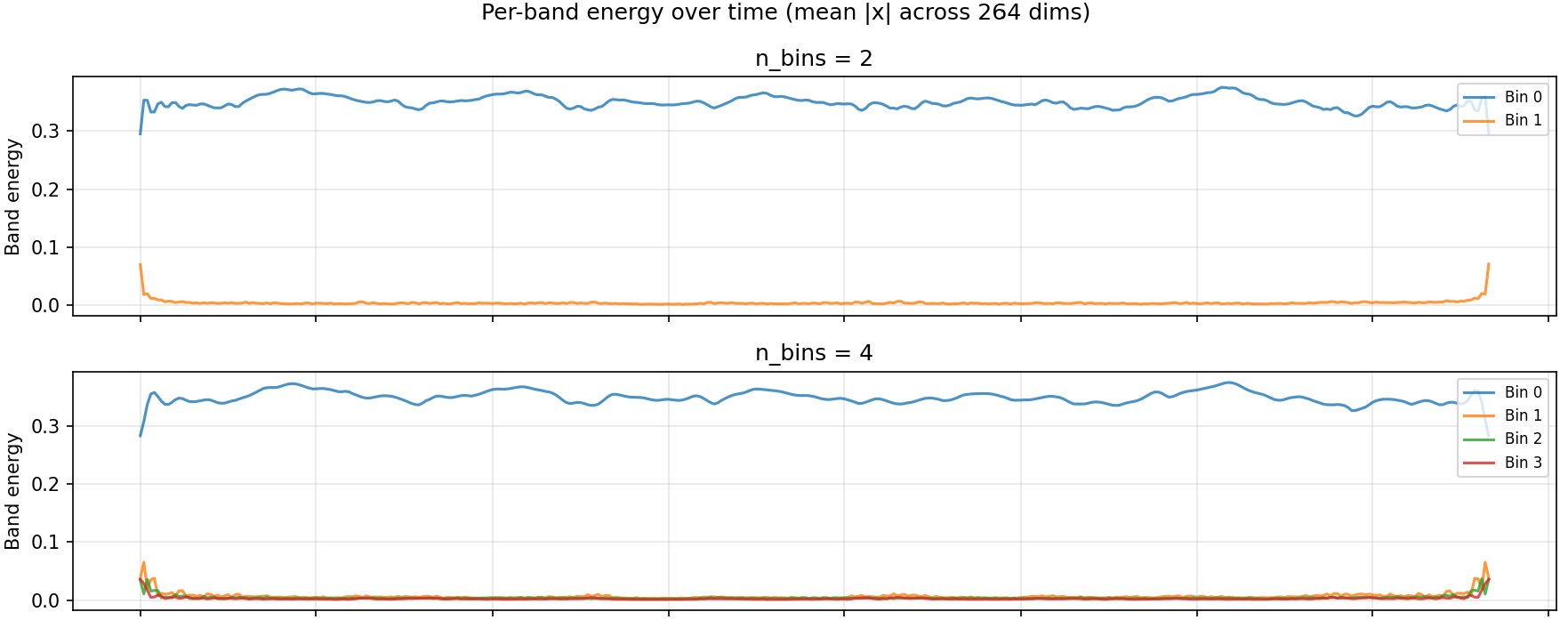}
\caption{Frequency energy distribution of reference dance motions. Most energy is concentrated in the low-frequency band, justifying the choice of $\text{nbins}=2$.}
\label{fig:freq_bands}
\end{figure}

We study the effect of the number of frequency bands (\texttt{nbins}) used in Cadence-MoE, conducting experiments on a 60\% subset of the training data. As shown in Table~\ref{tab:nbins}, $\text{nbins}=2$ outperforms $\text{nbins}=4$.
As illustrated in Fig.~\ref{fig:freq_bands}, most informative motion dynamics are concentrated in the low-frequency band; overly fine-grained frequency partitioning disperses useful information and weakens cadence modeling, supporting our design choice of two frequency bands.

\begin{table}[H]
\caption{Cadence-MoE frequency band ablation (60\% training data subset).}
\label{tab:nbins}
\centering
\resizebox{0.62\linewidth}{!}{
\begin{tabular}{lccccc}
\hline
nbins & FID$_k$ $\downarrow$ & FID$_m$ $\downarrow$ & Div$_k$ $\uparrow$ & Div$_m$ $\uparrow$ & BAS $\uparrow$ \\
\hline
2 & \textbf{43.19} & \textbf{18.38} & 9.05 & 7.03 & 0.2335 \\
4 & 118.49 & 58.47 & \textbf{12.95} & \textbf{12.44} & \textbf{0.2375} \\
\hline
\end{tabular}
}
\end{table}

\section*{C.4 Additional Comparisons and Computational Cost}

\noindent\textbf{FineNet Comparison}
We additionally compare against FineNet~\cite{li2023finedance} following the evaluation protocol in the main paper (Table~\ref{tab:sotacompare}). FineNet achieves FID$_k$ = 94.39, FSR = 13.53\%, Div$_k$ = 4.42, and BAS = 0.2318, substantially lower than our \MethodName\ (FID$_k$ = \textbf{32.37}, FSR = \textbf{5.33\%}, Div$_k$ = \textbf{7.34}, BAS = \textbf{0.2342}).

\noindent\textbf{Computational Cost}
Training the full \MethodName\ model on 4$\times$A100 GPUs requires approximately 18 hours; removing RAG and MoE reduces training time to 14 hours.
At inference, \MethodName\ runs at \textbf{42.8 FPS} on a single A100 GPU.
The RAG module incurs a 34.2\% overhead and the MoE module a 12.0\% overhead, both manageable due to offline feature extraction and parallel retrieval.

\section*{C.5 Addition Visualization Results}

\begin{figure}[!h]
  \centering
  \includegraphics[width=0.8\linewidth]{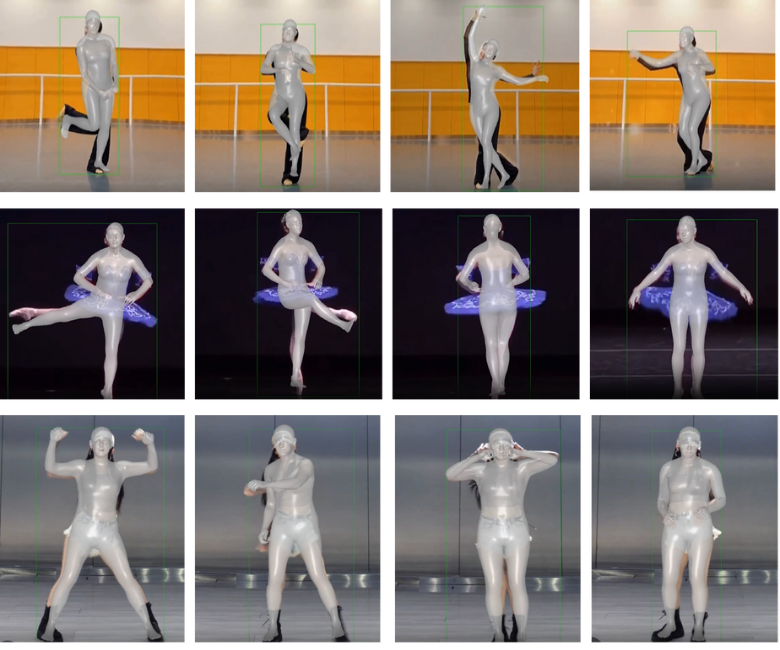}
\caption{Visualization of 3D motion capture results from monocular video using GVHMR and SMPLest-X. While this step already captures fine-grained motion details, the results still suffer from physically implausible artifacts. Therefore, we do not use these motions directly. Instead, we further refine them with a physical simulation-based correction and our Foot Restoration Diffusion Model (FRDM) to enforce accurate foot–ground contact.}
\label{fig:GVHMR}
\end{figure}

\noindent\textbf{Visualization Results of Our 3D Motion Capture.}
We use GVHMR and SMPLest-X to reconstruct 3D full-body motion from monocular dance videos, capturing hand gestures and facial expressions. Fig.~\ref{fig:GVHMR} shows results for diverse dances.

\begin{figure}[!h]
  \centering
  \includegraphics[width=0.8\linewidth]{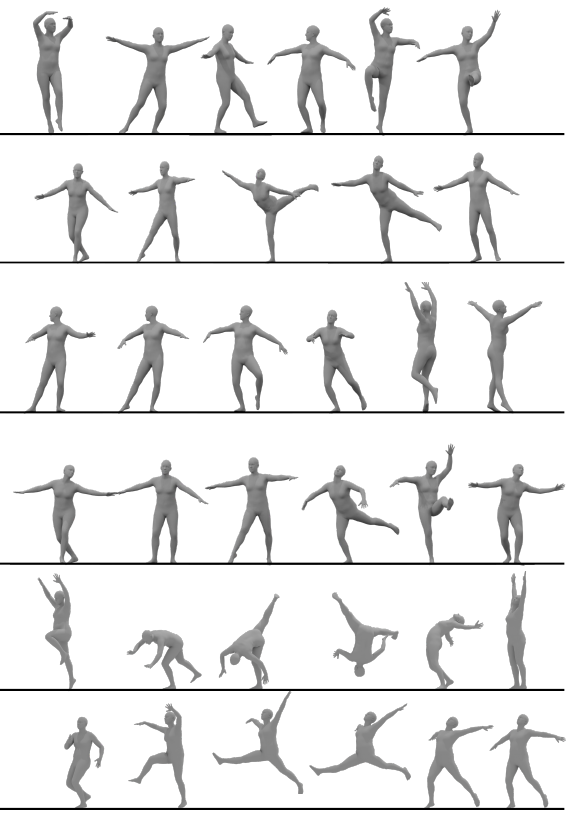}
\caption{Visualization of InfiniteDance.}
\label{fig:INfinitedata_show3D}
\end{figure}

\clearpage 
\noindent\textbf{Visualization of InfiniteDance.}
Fig.~\ref{fig:INfinitedata_show3D} illustrates motion sequences sampled from our {InfiniteDance} dataset, which features high-quality motion without artifacts like foot sliding or penetration. It includes both common dance moves and challenging actions such as spins, flips, and jumps.

\section*{D. 2D Dance Video Generation}

\subsection*{D.1. Task}
Pose-guided 2D human animation video generation has seen significant advancements, particularly with
the incorporation of advanced pose estimation methods and powerful generative diffusion models.
This task aims to synthesize animated video sequences utilizing a reference character image and a sequence of motion signals, which can be formulated as learning a function $V_{1:N} = F_{im2v}(I_{ref}, M_{1:N})$, where $I_{ref}$ is reference image and $M_{1:N}$ is motion sequence.
To better support this task, our dataset provides a variety of motion signals, including dwpose, depth maps, normal maps, and dense pose. Furthermore, we expand the diversity of dance genres, which addresses a limitation of existing datasets.
\subsection*{D.2. Method}

We utilize the advanced video diffusion transformer model, wan2.1~\cite{wan2025}, as the backbone for human animation. Our objective is to enable video generation models to synthesize human dance videos that retain the subject’s identity and appearance while producing accurate and natural motion aligned with given signals. However, fine-tuning a pretrained model on new data often leads to overfitting and degradation of previously learned general knowledge.

To address this, we adopt the Low-Rank Adaptation (LoRA~\cite{hu2022lora}) method, which introduces a small number of trainable parameters while freezing the original weights. This allows efficient adaptation to new tasks without compromising the model’s generalization ability. Furthermore, we design an auxiliary motion guidance module consisting of 3D convolutional layers to extract spatiotemporal features from motion signals. These features are injected into the model by adding them to the patchified noise input, guiding the diffusion process in a motion-aware manner.
In addition, we incorporate an auxiliary motion guidance module to enable the injection of motion signals. This module consists of multiple 3D convolutional layers designed to extract spatiotemporal features. The extracted features are then added to the patchified noise and subsequently fed into the baseline DiT.

\subsection*{D.3. Expriments}

\begin{figure}[t!]
  \centering
  \includegraphics[width=0.88\linewidth]{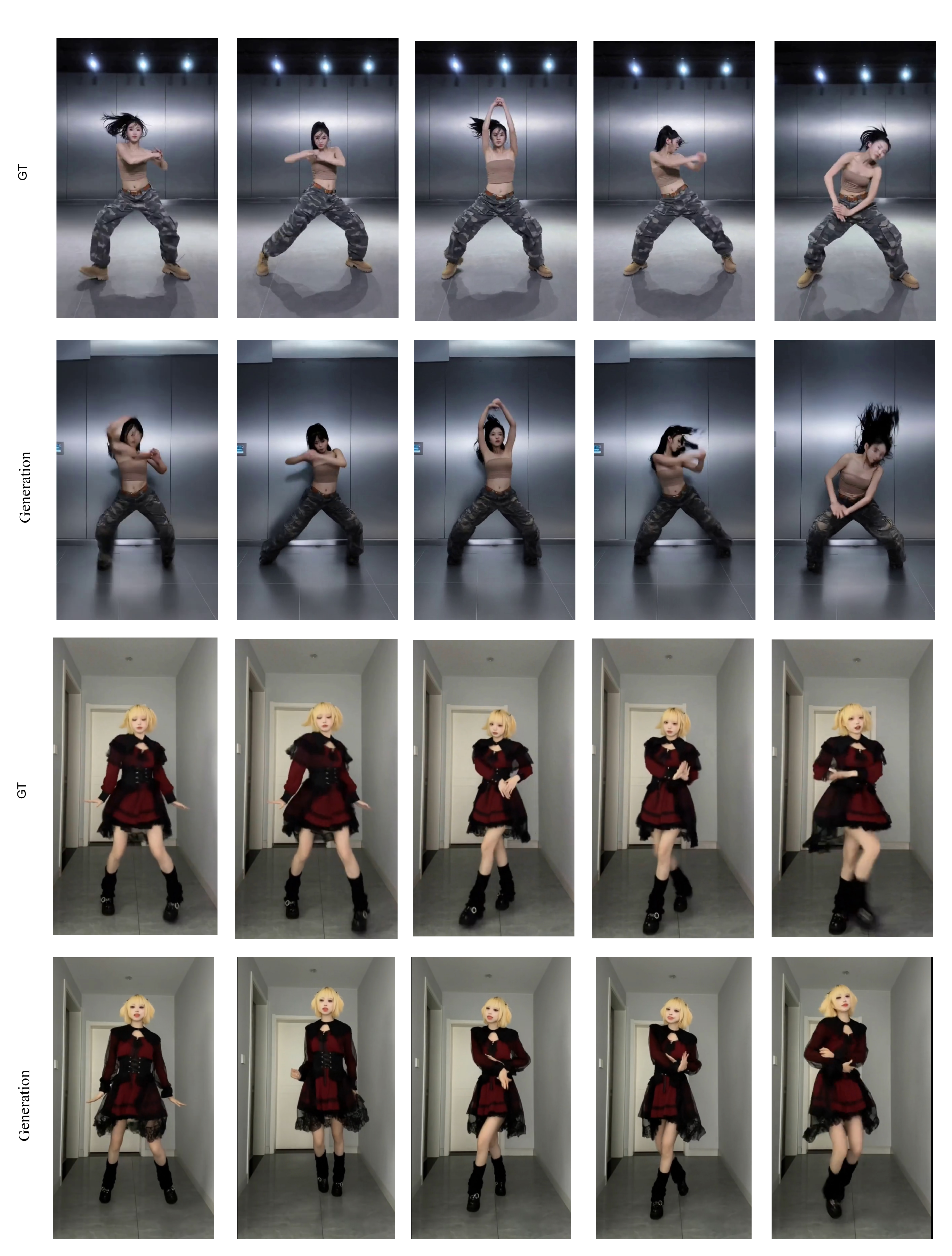}
\caption{Visualization of pose-guided 2D dance generation}
\label{fig:INfinitedata_show}
\end{figure}

We compare our approach against state-of-the-art
pose-guided human video generation methods, including Disco \cite{wang2023disco}, Moore-AnimateAnyone \cite{hu2024animate},
and Champ \cite{zhu2024champ}. 
For evaluation metrics, we evaluate image quality using L1 error, Peak Signal-to-Noise
Ratio (PSNR), Structural Similarity Index Measure (SSIM), Learned Perceptual Image Patch Similarity (LPIPS), and Frechet Inception Distance (FID). In addition, video-level FID (FID-VID) and Frechet Video Distance (FVD) are employed to evaluate the quality of the generated videos.Additionally, visualization results are presented in Fig.~\ref{fig:INfinitedata_show}.

\begin{table}[th]
\caption{Quantitative comparison with SOTA methods.}
\centering
\resizebox{0.78\textwidth}{!}{
	\begin{tabular}{lccccccc}
		\toprule [1pt] 
		Method & $\mathrm{FID}\downarrow$ & $\mathrm{SSIM}\uparrow$ & $\mathrm{PSNR}\uparrow$ & $\mathrm{LPIPS}\downarrow$ & L1 $\downarrow$ & FID-FVD $\downarrow$ & FID $\downarrow$\\
        \noalign{\smallskip}\hline\noalign{\smallskip}
        DisCo \cite{wang2023disco} & 57.84 & 0.51 & 10.66 & 0.47 & 2.4e-4 & 43.17 & 522.28\\
        Animate Anyone \cite{hu2024animate} & 45.90 & 0.54 & 12.82 & 0.44 & 1.5e-4 & 33.32 & 505.01 \\
        Champ \cite{zhu2024champ} & 42.61 & 0.57 & 13.21 & 0.41 &1.3e-4 & 31.27 & 447.21 \\
        \noalign{\smallskip}\hline\noalign{\smallskip}
        Ours & \textbf{37.94} & \textbf{0.62} & \textbf{14.60} & \textbf{0.37} & \textbf{9.7e-5} & \textbf{24.34} &\textbf{405.27}\\
		\bottomrule [1pt] 
	\end{tabular}
}
\label{tab:human_animation}   
\end{table}

%% file: main.bib
@String(ICCV  = {Int. Conf. Comput. Vis.})

@String(ICLR  = {Int. Conf. Learn. Represent.})

@String(AAAI  = {AAAI})

@String(ICASSP=	{ICASSP})

@String(TOG   = {ACM Trans. Graph.})

@String(ICCV  = {ICCV})

@String(ICLR  = {ICLR})

@String(TOG   = {ACM TOG})

@inproceedings{humanml3d,
  title={Generating diverse and natural 3d human motions from text},
  author={Guo, Chuan and Zou, Shihao and Zuo, Xinxin and Wang, Sen and Ji, Wei and Li, Xingyu and Cheng, Li},
  booktitle={Proceedings of the IEEE/CVF conference on computer vision and pattern recognition},
  pages={5152--5161},
  year={2022}
}

@ArtifactSoftware{R,
    title = {R: A Language and Environment for Statistical Computing},
    author = {{R Core Team}},
    organization = {R Foundation for Statistical Computing},
    address = {Vienna, Austria},
    year = {2019},
    url = {https://www.R-project.org/},
}

@article{motorica2024,
  title={Listen, Denoise, Action! Audio-Driven Motion Synthesis with Diffusion Models},
  author={Alexanderson, Simon and Nagy, Rajmund and Beskow, Jonas and Henter, Gustav Eje},
  year={2023},
  issue_date={August 2023},
  publisher={ACM},
  volume={42},
  number={4},
  doi={10.1145/3592458},
  journal={ACM Trans. Graph.},
  articleno={44},
  numpages={20},
  pages={44:1--44:20}
}

@misc{li2023finedance,
      title={FineDance: A Fine-grained Choreography Dataset for 3D Full Body Dance Generation}, 
      author={Ronghui Li and Junfan Zhao and Yachao Zhang and Mingyang Su and Zeping Ren and Han Zhang and Yansong Tang and Xiu Li},
      year={2023},
      eprint={2212.03741},
      archivePrefix={arXiv},
      primaryClass={cs.CV},
      url={https://arxiv.org/abs/2212.03741}, 
}

@article{ling2024motionllama,
  title={MotionLLaMA: A Unified Framework for Motion Synthesis and Comprehension},
  author={Ling, Zeyu and Han, Bo and Li, Shiyang and Shen, Hongdeng and Cheng, Jikang and Zou, Changqing},
  journal={arXiv preprint arXiv:2411.17335},
  year={2024}
}

@inproceedings{guo2024momask,
  title={Momask: Generative masked modeling of 3d human motions},
  author={Guo, Chuan and Mu, Yuxuan and Javed, Muhammad Gohar and Wang, Sen and Cheng, Li},
  booktitle={Proceedings of the IEEE/CVF Conference on Computer Vision and Pattern Recognition},
  pages={1900--1910},
  year={2024}
}

@inproceedings{li2021hybrik,
  title={Hybrik: A hybrid analytical-neural inverse kinematics solution for 3d human pose and shape estimation},
  author={Li, Jiefeng and Xu, Chao and Chen, Zhicun and Bian, Siyuan and Yang, Lixin and Lu, Cewu},
  booktitle={Proceedings of the IEEE/CVF conference on computer vision and pattern recognition},
  pages={3383--3393},
  year={2021}
}

@article{li2025hybrik,
  title={Hybrik-x: Hybrid analytical-neural inverse kinematics for whole-body mesh recovery},
  author={Li, Jiefeng and Bian, Siyuan and Xu, Chao and Chen, Zhicun and Yang, Lixin and Lu, Cewu},
  journal={IEEE Transactions on Pattern Analysis and Machine Intelligence},
  year={2025},
  publisher={IEEE}
}

@article{danceformer,
author = {Li, Buyu and Zhao, Yongchi and Zhelun, Shi and Sheng, Lu},
year = {2022},
month = {06},
pages = {1272-1279},
title = {DanceFormer: Music Conditioned 3D Dance Generation with Parametric Motion Transformer},
volume = {36},
journal = {Proceedings of the AAAI Conference on Artificial Intelligence},
doi = {10.1609/aaai.v36i2.20014}
}

@inproceedings{dancewithmelody,
  author    = {Tang, Taoran and Jia, Jia and Hanyang, Mao},
  title     = {Dance with Melody: An LSTM-autoencoder Approach to Music-oriented Dance Synthesis},
  booktitle = {ACM International Conference on Multimedia},
  year      = {2018},
  pages     = {1598-1606},
  doi       = {10.1145/3240508.3240526}
}

@article{ChoreMaster,
  title     = {ChoreoMaster: Choreography-Oriented Music-Driven Dance Synthesis},
  author    = {Chen, Kang and Tan, Zhipeng and Lei, Jin and Zhang, Song-Hai and Guo, Yuan-Chen and Zhang, Weidong and Hu, Shi-Min},
  journal   = {ACM Transactions on Graphics (TOG)},
  year      = {2021},
  volume    = {40},
  number    = {4},
  pages     = {1--13}
}

@misc{luo2024popdgpopular3ddance,
      title={POPDG: Popular 3D Dance Generation with PopDanceSet}, 
      author={Zhenye Luo and Min Ren and Xuecai Hu and Yongzhen Huang and Li Yao},
      year={2024},
      eprint={2405.03178},
      archivePrefix={arXiv},
      primaryClass={cs.SD},
      url={https://arxiv.org/abs/2405.03178}, 
}

@ARTICLE{deepdance,  author={Sun, Guofei and Wong, Yongkang and Cheng, Zhiyong and Kankanhalli, Mohan S. and Geng, Weidong and Li, Xiangdong},  journal={IEEE Transactions on Multimedia},   title={DeepDance: Music-to-Dance Motion Choreography With Adversarial Learning},   year={2021},  volume={23},  number={},  pages={497-509},  doi={10.1109/TMM.2020.2981989}}

@misc{wang2024dancecamera3d3dcameramovement,
      title={DanceCamera3D: 3D Camera Movement Synthesis with Music and Dance}, 
      author={Zixuan Wang and Jia Jia and Shikun Sun and Haozhe Wu and Rong Han and Zhenyu Li and Di Tang and Jiaqing Zhou and Jiebo Luo},
      year={2024},
      eprint={2403.13667},
      archivePrefix={arXiv},
      primaryClass={cs.CV},
      url={https://arxiv.org/abs/2403.13667}, 
}

@misc{siyao2024duolandofollowergptoffpolicy,
      title={Duolando: Follower GPT with Off-Policy Reinforcement Learning for Dance Accompaniment}, 
      author={Li Siyao and Tianpei Gu and Zhitao Yang and Zhengyu Lin and Ziwei Liu and Henghui Ding and Lei Yang and Chen Change Loy},
      year={2024},
      eprint={2403.18811},
      archivePrefix={arXiv},
      primaryClass={cs.CV},
      url={https://arxiv.org/abs/2403.18811}, 
}

@misc{li2024interdancereactive3ddancegeneration,
      title={InterDance:Reactive 3D Dance Generation with Realistic Duet Interactions}, 
      author={Ronghui Li and Youliang Zhang and Yachao Zhang and Yuxiang Zhang and Mingyang Su and Jie Guo and Ziwei Liu and Yebin Liu and Xiu Li},
      year={2024},
      eprint={2412.16982},
      archivePrefix={arXiv},
      primaryClass={cs.CV},
      url={https://arxiv.org/abs/2412.16982}, 
}

@misc{le2023musicdrivengroupchoreography,
      title={Music-Driven Group Choreography}, 
      author={Nhat Le and Thang Pham and Tuong Do and Erman Tjiputra and Quang D. Tran and Anh Nguyen},
      year={2023},
      eprint={2303.12337},
      archivePrefix={arXiv},
      primaryClass={cs.MM},
      url={https://arxiv.org/abs/2303.12337}, 
}

@inproceedings{choreographers_motiongraph,
  title={How do choreographers craft dance? Designing for a choreographer-technology partnership},
  author={Ciolfi Felice, Marianela and Alaoui, Sarah Fdili and Mackay, Wendy E},
  booktitle={Proceedings of the 3rd International Symposium on Movement and Computing},
  pages={1--8},
  year={2016}
}

@inproceedings{berman2015kinetic,
  title={Kinetic imaginations: exploring the possibilities of combining AI and dance},
  author={Berman, Alexander and James, Valencia},
  booktitle={Twenty-Fourth International Joint Conference on Artificial Intelligence},
  pages={2431–2437},
  year={2015}
}

@article{ofli2011learn2dance,
  title={Learn2dance: Learning statistical music-to-dance mappings for choreography synthesis},
  author={Ofli, Ferda and Erzin, Engin and Yemez, Y{\"u}cel and Tekalp, A Murat},
  journal={IEEE Transactions on Multimedia},
  volume={14},
  number={3},
  pages={747--759},
  year={2011},
  publisher={IEEE}
}

@inproceedings{kim2022brand_sequencemodel_based,
  title={A brand new dance partner: Music-conditioned pluralistic dancing controlled by multiple dance genres},
  author={Kim, Jinwoo and Oh, Heeseok and Kim, Seongjean and Tong, Hoseok and Lee, Sanghoon},
  booktitle={Proceedings of the IEEE/CVF Conference on Computer Vision and Pattern Recognition},
  pages={3490--3500},
  year={2022}
}

@inproceedings{aist++,
  title={Ai choreographer: Music conditioned 3d dance generation with aist++},
  author={Li, Ruilong and Yang, Shan and Ross, David A and Kanazawa, Angjoo},
  booktitle={Proceedings of the IEEE/CVF International Conference on Computer Vision},
  pages={13401--13412},
  year={2021}
}

@article{luo2024m3gpt,
  title={M3 GPT: An Advanced Multimodal, Multitask Framework for Motion Comprehension and Generation},
  author={Luo, Mingshuang and Hou, Ruibing and Li, Zhuo and Chang, Hong and Liu, Zimo and Wang, Yaowei and Shan, Shiguang},
  journal={arXiv preprint arXiv:2405.16273},
  year={2024}
}

@article{zhang2025motion,
  title={Motion Anything: Any to Motion Generation},
  author={Zhang, Zeyu and Wang, Yiran and Mao, Wei and Li, Danning and Zhao, Rui and Wu, Biao and Song, Zirui and Zhuang, Bohan and Reid, Ian and Hartley, Richard},
  journal={arXiv preprint arXiv:2503.06955},
  year={2025}
}

@article{wang2023disco,
  title={Disco: Disentangled control for referring human dance generation in real world},
  author={Wang, Tan and Li, Linjie and Lin, Kevin and Lin, Chung-Ching and Yang, Zhengyuan and Zhang, Hanwang and Liu, Zicheng and Wang, Lijuan},
  journal={arXiv preprint arXiv:2307.00040},
  volume={2},
  number={3},
  pages={4},
  year={2023}
}

@inproceedings{hu2024animate,
  title={Animate anyone: Consistent and controllable image-to-video synthesis for character animation},
  author={Hu, Li},
  booktitle={Proceedings of the IEEE/CVF Conference on Computer Vision and Pattern Recognition},
  pages={8153--8163},
  year={2024}
}

@inproceedings{zhu2024champ,
  title={Champ: Controllable and consistent human image animation with 3d parametric guidance},
  author={Zhu, Shenhao and Chen, Junming Leo and Dai, Zuozhuo and Dong, Zilong and Xu, Yinghui and Cao, Xun and Yao, Yao and Zhu, Hao and Zhu, Siyu},
  booktitle={European Conference on Computer Vision},
  pages={145--162},
  year={2024},
  organization={Springer}
}

@inproceedings{siyao2022bailando,
  title={Bailando: 3D Dance Generation by Actor-Critic GPT with Choreographic Memory},
  author={Siyao, Li and Yu, Weijiang and Gu, Tianpei and Lin, Chunze and Wang, Quan and Qian, Chen and Loy, Chen Change and Liu, Ziwei},
  booktitle={Proceedings of the IEEE/CVF Conference on Computer Vision and Pattern Recognition},
  pages={11050--11059},
  year={2022}
}

@inproceedings{gtn,
  title={GTN-Bailando: Genre Consistent long-Term 3D Dance Generation Based on Pre-Trained Genre Token Network},
  author={Zhuang, Haolin and Lei, Shun and Xiao, Long and Li, Weiqin and Chen, Liyang and Yang, Sicheng and Wu, Zhiyong and Kang, Shiyin and Meng, Helen},
  booktitle={ICASSP 2023-2023 IEEE International Conference on Acoustics, Speech and Signal Processing (ICASSP)},
  pages={1--5},
  year={2023},
  organization={IEEE}
}

@inproceedings{edge,
  title={Edge: Editable dance generation from music},
  author={Tseng, Jonathan and Castellon, Rodrigo and Liu, Karen},
  booktitle={Proceedings of the IEEE/CVF Conference on Computer Vision and Pattern Recognition},
  pages={448--458},
  year={2023}
}

@misc{li2024lodge,
      title={Lodge: A Coarse to Fine Diffusion Network for Long Dance Generation Guided by the Characteristic Dance Primitives}, 
      author={Ronghui Li and YuXiang Zhang and Yachao Zhang and Hongwen Zhang and Jie Guo and Yan Zhang and Yebin Liu and Xiu Li},
      year={2024},
      eprint={2403.10518},
      archivePrefix={arXiv},
      primaryClass={cs.CV},
      url={https://arxiv.org/abs/2403.10518}, 
}

@incollection{loper2023smpl,
  title={SMPL: A skinned multi-person linear model},
  author={Loper, Matthew and Mahmood, Naureen and Romero, Javier and Pons-Moll, Gerard and Black, Michael J},
  booktitle={Seminal Graphics Papers: Pushing the Boundaries, Volume 2},
  pages={851--866},
  year={2023}
}

@inproceedings{shen2024world,
  title={World-Grounded Human Motion Recovery via Gravity-View Coordinates},
  author={Shen, Zehong and Pi, Huaijin and Xia, Yan and Cen, Zhi and Peng, Sida and Hu, Zechen and Bao, Hujun and Hu, Ruizhen and Zhou, Xiaowei},
  booktitle={SIGGRAPH Asia 2024 Conference Papers},
  pages={1--11},
  year={2024}
}

@article{li2024lodge++,
  title={Lodge++: High-quality and long dance generation with vivid choreography patterns},
  author={Li, Ronghui and Zhang, Hongwen and Zhang, Yachao and Zhang, Yuxiang and Zhang, Youliang and Guo, Jie and Zhang, Yan and Li, Xiu and Liu, Yebin},
  journal={arXiv preprint arXiv:2410.20389},
  year={2024}
}

@inproceedings{varghese2024yolov8,
  title={Yolov8: A novel object detection algorithm with enhanced performance and robustness},
  author={Varghese, Rejin and Sambath, M},
  booktitle={2024 International Conference on Advances in Data Engineering and Intelligent Computing Systems (ADICS)},
  pages={1--6},
  year={2024},
  organization={IEEE}
}

@article{yin2025smplest,
  title={SMPLest-X: Ultimate Scaling for Expressive Human Pose and Shape Estimation},
  author={Yin, Wanqi and Cai, Zhongang and Wang, Ruisi and Zeng, Ailing and Wei, Chen and Sun, Qingping and Mei, Haiyi and Wang, Yanjun and Pang, Hui En and Zhang, Mingyuan and others},
  journal={arXiv preprint arXiv:2501.09782},
  year={2025}
}

@article{zhang2024plug,
  title={A Plug-and-Play Physical Motion Restoration Approach for In-the-Wild High-Difficulty Motions},
  author={Zhang, Youliang and Li, Ronghui and Zhang, Yachao and Pan, Liang and Wang, Jingbo and Liu, Yebin and Li, Xiu},
  journal={arXiv preprint arXiv:2412.17377},
  year={2024}
}

@inproceedings{Luo2023PerpetualHC,
    author={Zhengyi Luo and Jinkun Cao and Alexander W. Winkler and Kris Kitani and Weipeng Xu},
    title={Perpetual Humanoid Control for Real-time Simulated Avatars},
    booktitle={International Conference on Computer Vision (ICCV)},
    year={2023}
}

@inproceedings{erez2015simulation,
  title={Simulation tools for model-based robotics: Comparison of bullet, havok, mujoco, ode and physx},
  author={Erez, Tom and Tassa, Yuval and Todorov, Emanuel},
  booktitle={2015 IEEE international conference on robotics and automation (ICRA)},
  pages={4397--4404},
  year={2015},
  organization={IEEE}
}

@article{hu2022lora,
  title={Lora: Low-rank adaptation of large language models.},
  author={Hu, Edward J and Shen, Yelong and Wallis, Phillip and Allen-Zhu, Zeyuan and Li, Yuanzhi and Wang, Shean and Wang, Lu and Chen, Weizhu and others},
  journal={ICLR},
  volume={1},
  number={2},
  pages={3},
  year={2022}
}

@article{wan2025,
      title={Wan: Open and Advanced Large-Scale Video Generative Models}, 
      author={Team Wan and Ang Wang and Baole Ai and Bin Wen and Chaojie Mao and Chen-Wei Xie and Di Chen and Feiwu Yu and Haiming Zhao and Jianxiao Yang and Jianyuan Zeng and Jiayu Wang and Jingfeng Zhang and Jingren Zhou and Jinkai Wang and Jixuan Chen and Kai Zhu and Kang Zhao and Keyu Yan and Lianghua Huang and Mengyang Feng and Ningyi Zhang and Pandeng Li and Pingyu Wu and Ruihang Chu and Ruili Feng and Shiwei Zhang and Siyang Sun and Tao Fang and Tianxing Wang and Tianyi Gui and Tingyu Weng and Tong Shen and Wei Lin and Wei Wang and Wei Wang and Wenmeng Zhou and Wente Wang and Wenting Shen and Wenyuan Yu and Xianzhong Shi and Xiaoming Huang and Xin Xu and Yan Kou and Yangyu Lv and Yifei Li and Yijing Liu and Yiming Wang and Yingya Zhang and Yitong Huang and Yong Li and You Wu and Yu Liu and Yulin Pan and Yun Zheng and Yuntao Hong and Yupeng Shi and Yutong Feng and Zeyinzi Jiang and Zhen Han and Zhi-Fan Wu and Ziyu Liu},
      journal = {arXiv preprint arXiv:2503.20314},
      year={2025}
}

@article{infonce,
  title={Representation learning with contrastive predictive coding},
  author={Oord, Aaron van den and Li, Yazhe and Vinyals, Oriol},
  journal={arXiv preprint arXiv:1807.03748},
  year={2018}
}

@inproceedings{shen2021efficient,
  title={Efficient attention: Attention with linear complexities},
  author={Shen, Zhuoran and Zhang, Mingyuan and Zhao, Haiyu and Yi, Shuai and Li, Hongsheng},
  booktitle={Proceedings of the IEEE/CVF winter conference on applications of computer vision},
  pages={3531--3539},
  year={2021}
}

@article{makoviychuk2021isaac,
  title={Isaac gym: High performance gpu-based physics simulation for robot learning},
  author={Makoviychuk, Viktor and Wawrzyniak, Lukasz and Guo, Yunrong and Lu, Michelle and Storey, Kier and Macklin, Miles and Hoeller, David and Rudin, Nikita and Allshire, Arthur and Handa, Ankur and others},
  journal={arXiv preprint arXiv:2108.10470},
  year={2021}
}

@inproceedings{radford2021learning,
  title={Learning transferable visual models from natural language supervision},
  author={Radford, Alec and Kim, Jong Wook and Hallacy, Chris and Ramesh, Aditya and Goh, Gabriel and Agarwal, Sandhini and Sastry, Girish and Askell, Amanda and Mishkin, Pamela and Clark, Jack and others},
  booktitle={International conference on machine learning},
  pages={8748--8763},
  year={2021},
  organization={PmLR}
}

@article{music2dance,
  title={Music2dance: Dancenet for music-driven dance generation},
  author={Zhuang, Wenlin and Wang, Congyi and Chai, Jinxiang and Wang, Yangang and Shao, Ming and Xia, Siyu},
  journal={ACM Transactions on Multimedia Computing, Communications, and Applications (TOMM)},
  volume={18},
  number={2},
  pages={1--21},
  year={2022},
  publisher={ACM New York, NY}
}

@misc{motiongpt,
      title={MotionGPT: Human Motion as a Foreign Language}, 
      author={Biao Jiang and Xin Chen and Wen Liu and Jingyi Yu and Gang Yu and Tao Chen},
      year={2023},
      eprint={2306.14795},
      archivePrefix={arXiv},
      primaryClass={cs.CV},
      url={https://arxiv.org/abs/2306.14795}, 
}

@misc{t2mgpt,
      title={T2M-GPT: Generating Human Motion from Textual Descriptions with Discrete Representations}, 
      author={Jianrong Zhang and Yangsong Zhang and Xiaodong Cun and Shaoli Huang and Yong Zhang and Hongwei Zhao and Hongtao Lu and Xi Shen},
      year={2023},
      eprint={2301.06052},
      archivePrefix={arXiv},
      primaryClass={cs.CV},
      url={https://arxiv.org/abs/2301.06052}, 
}

@misc{heusel2018ganstrainedtimescaleupdate,
      title={GANs Trained by a Two Time-Scale Update Rule Converge to a Local Nash Equilibrium}, 
      author={Martin Heusel and Hubert Ramsauer and Thomas Unterthiner and Bernhard Nessler and Sepp Hochreiter},
      year={2018},
      eprint={1706.08500},
      archivePrefix={arXiv},
      primaryClass={cs.LG},
      url={https://arxiv.org/abs/1706.08500}, 
}

@inproceedings{Sun_2024, series={SIGGRAPH ’24},
   title={LGTM: Local-to-Global Text-Driven Human Motion Diffusion Model},
   url={http://dx.doi.org/10.1145/3641519.3657422},
   DOI={10.1145/3641519.3657422},
   booktitle={Special Interest Group on Computer Graphics and Interactive Techniques Conference Conference Papers ’24},
   publisher={ACM},
   author={Sun, Haowen and Zheng, Ruikun and Huang, Haibin and Ma, Chongyang and Huang, Hui and Hu, Ruizhen},
   year={2024},
   month=jul, pages={1–9},
   collection={SIGGRAPH ’24} }

@misc{zhang2023teditemporallyentangleddiffusionlongterm,
      title={TEDi: Temporally-Entangled Diffusion for Long-Term Motion Synthesis}, 
      author={Zihan Zhang and Richard Liu and Kfir Aberman and Rana Hanocka},
      year={2023},
      eprint={2307.15042},
      archivePrefix={arXiv},
      primaryClass={cs.CV},
      url={https://arxiv.org/abs/2307.15042}, 
}

@inproceedings{Goel_2024, series={SIGGRAPH ’24},
   title={Iterative Motion Editing with Natural Language},
   url={http://dx.doi.org/10.1145/3641519.3657447},
   DOI={10.1145/3641519.3657447},
   booktitle={Special Interest Group on Computer Graphics and Interactive Techniques Conference Conference Papers ’24},
   publisher={ACM},
   author={Goel, Purvi and Wang, Kuan-Chieh and Liu, C. Karen and Fatahalian, Kayvon},
   year={2024},
   month=jul, pages={1–9},
   collection={SIGGRAPH ’24} }

@misc{cohan2024flexiblemotioninbetweeningdiffusion,
      title={Flexible Motion In-betweening with Diffusion Models}, 
      author={Setareh Cohan and Guy Tevet and Daniele Reda and Xue Bin Peng and Michiel van de Panne},
      year={2024},
      eprint={2405.11126},
      archivePrefix={arXiv},
      primaryClass={cs.CV},
      url={https://arxiv.org/abs/2405.11126}, 
}

@misc{luo2023perpetualhumanoidcontrolrealtime,
      title={Perpetual Humanoid Control for Real-time Simulated Avatars}, 
      author={Zhengyi Luo and Jinkun Cao and Alexander Winkler and Kris Kitani and Weipeng Xu},
      year={2023},
      eprint={2305.06456},
      archivePrefix={arXiv},
      primaryClass={cs.CV},
      url={https://arxiv.org/abs/2305.06456}, 
}

@misc{touvron2023llamaopenefficientfoundation,
      title={LLaMA: Open and Efficient Foundation Language Models}, 
      author={Hugo Touvron and Thibaut Lavril and Gautier Izacard and Xavier Martinet and Marie-Anne Lachaux and Timothée Lacroix and Baptiste Rozière and Naman Goyal and Eric Hambro and Faisal Azhar and Aurelien Rodriguez and Armand Joulin and Edouard Grave and Guillaume Lample},
      year={2023},
      eprint={2302.13971},
      archivePrefix={arXiv},
      primaryClass={cs.CL},
      url={https://arxiv.org/abs/2302.13971}, 
}

@misc{touvron2023llama2openfoundation,
      title={Llama 2: Open Foundation and Fine-Tuned Chat Models}, 
      author={Hugo Touvron and Louis Martin and Kevin Stone and Peter Albert and Amjad Almahairi and Yasmine Babaei and Nikolay Bashlykov and Soumya Batra and Prajjwal Bhargava and Shruti Bhosale and Dan Bikel and Lukas Blecher and Cristian Canton Ferrer and Moya Chen and Guillem Cucurull and David Esiobu and Jude Fernandes and Jeremy Fu and Wenyin Fu and Brian Fuller and Cynthia Gao and Vedanuj Goswami and Naman Goyal and Anthony Hartshorn and Saghar Hosseini and Rui Hou and Hakan Inan and Marcin Kardas and Viktor Kerkez and Madian Khabsa and Isabel Kloumann and Artem Korenev and Punit Singh Koura and Marie-Anne Lachaux and Thibaut Lavril and Jenya Lee and Diana Liskovich and Yinghai Lu and Yuning Mao and Xavier Martinet and Todor Mihaylov and Pushkar Mishra and Igor Molybog and Yixin Nie and Andrew Poulton and Jeremy Reizenstein and Rashi Rungta and Kalyan Saladi and Alan Schelten and Ruan Silva and Eric Michael Smith and Ranjan Subramanian and Xiaoqing Ellen Tan and Binh Tang and Ross Taylor and Adina Williams and Jian Xiang Kuan and Puxin Xu and Zheng Yan and Iliyan Zarov and Yuchen Zhang and Angela Fan and Melanie Kambadur and Sharan Narang and Aurelien Rodriguez and Robert Stojnic and Sergey Edunov and Thomas Scialom},
      year={2023},
      eprint={2307.09288},
      archivePrefix={arXiv},
      primaryClass={cs.CL},
      url={https://arxiv.org/abs/2307.09288}, 
}

@misc{grattafiori2024llama3herdmodels,
  title={The Llama 3 Herd of Models}, 
  author={Aaron Grattafiori, Abhimanyu Dubey, et al.},
  year={2024},
  eprint={2407.21783},
  archivePrefix={arXiv},
  primaryClass={cs.AI},
  url={https://arxiv.org/abs/2407.21783},
}

@misc{zhu2025muqselfsupervisedmusicrepresentation,
      title={MuQ: Self-Supervised Music Representation Learning with Mel Residual Vector Quantization}, 
      author={Haina Zhu and Yizhi Zhou and Hangting Chen and Jianwei Yu and Ziyang Ma and Rongzhi Gu and Yi Luo and Wei Tan and Xie Chen},
      year={2025},
      eprint={2501.01108},
      archivePrefix={arXiv},
      primaryClass={cs.SD},
      url={https://arxiv.org/abs/2501.01108}, 
}
